\documentclass[conference,compsoc]{IEEEtran}

\ifCLASSOPTIONcompsoc
  \usepackage[nocompress]{cite}
\else
  \usepackage{cite}
\fi

\usepackage{algorithm}
\usepackage{algorithmic}

\usepackage{multirow}
\usepackage{makecell}
\usepackage{threeparttable}
\usepackage{adjustbox}

\usepackage{graphicx}
\usepackage{amsmath,amssymb,amsthm}
\usepackage{booktabs}
\usepackage{url}
\usepackage{microtype}
\usepackage[caption=false,font=footnotesize,labelfont=sf,textfont=sf]{subfig}

\begin{document}

\title{OPSD-Erase: On-Policy Concept Erasure for Text-to-Image Generative Models}

\author{
\IEEEauthorblockN{
Yunfei Zheng$^{1,2,*}$, 
Dandan Zheng$^{1,2,*}$, 
Laurence T. Yang$^{1}$, 
Shenghao Liu$^{1}$, 
Xiaoxuan Fan$^{1}$, 
Xianjun Deng$^{1,\dagger}$
}
\IEEEauthorblockA{
$^{1}$Huazhong University of Science and Technology, China
}
\IEEEauthorblockA{
$^{2}$Ant Group, China
}
\IEEEauthorblockA{
$^{*}$Equal contribution; $^{\dagger}$Corresponding author
}
}

\maketitle

\begin{abstract}
Open-weight text-to-image models can generate high-quality images from natural language prompts, but they can also be induced to produce unsafe content, creating substantial risks in open deployment. Existing defenses are insufficient for two reasons. External detection modules and inference-time guidance can be removed once the weights are released. More durable weight-editing methods, including closed-form editing and existing fine-tuning-based erasure, are typically trained on states that differ from those visited by the edited model during sampling, creating a training-inference inconsistency that leaves concept erasure incomplete. We address this problem by viewing text-to-image generation as a unified reverse process and defining concept erasure as alignment of the one-step reverse update on the states visited during sampling. We propose \textbf{OPSD-Erase}, an on-policy framework that trains on student-visited states, uses a frozen reference model to provide safe one-step targets, and uses anchor conditions to preserve benign generation behavior. A probe on the sampling trajectory reveals two mismatch patterns: under off-policy fixed-data supervision, update error grows as the rollout leaves the training support; more generally, non-on-policy methods retain substantial residual mismatch across visited states. In contrast, OPSD-Erase keeps the one-step reverse update aligned on the states it visits. Experiments on SD1.4 and Z-Image show that OPSD-Erase more reliably suppresses unsafe concept realization while preserving benign utility, and that this safety behavior remains robust under wording shift, multi-concept erasure, and sequential erasure.
\end{abstract}

\section{Introduction}
\label{sec:introduction}

Open-weight text-to-image generative models can synthesize high-quality images from natural-language prompts~\cite{Rombach_2022_CVPR,ICLR2024_081b0806,team2025zimage}, but the same capability also enables the generation of unsafe content at scale~\cite{wei2024exploring,gupta2022unstable,li2024art}. This creates a particularly challenging safety problem in open deployment: once model weights are released, users can run the model locally, remove external safeguards, and freely search for prompts that elicit prohibited content.

The real-world misuse is already substantial.
On Pixiv, Wei and Tyson find that the monthly share of restricted adult artworks rose from about 20\% to over 30\% after AI generation became available, and by December 2023 more than half of the R-18 artworks were AI-generated~\cite{wei2024understanding}.
On Civitai, a platform for open-source text-to-image models, De~Rosa~Palmini~\textit{et al.} record 6.5 million posted images from October 2023 to April 2024, and the share rated above PG-13 rises from 55.19\% to 72.94\%~\cite{palmini2024civiverse}.
Wagner and Cetini\'c extend the image record to 40.6 million images and find that the share flagged NSFW rises from 41\% in January 2023 to 80\% in December 2024~\cite{wagner2026perpetuating}.
These observations suggest that model-level safety should not be treated as a secondary deployment concern; for open-weight text-to-image systems, it is a core requirement.

A natural defense is to prevent the model from realizing an unsafe concept even when it is requested by the user. Existing approaches mainly fall into two categories.
The first category adds safety outside the released weights, such as prompt filtering, output detection, or inference-time guidance~\cite{markov2023holistic,yang2024guardt2i,liu2024latent,rando2022red,schramowski2022can,schramowski2023safe,yoon2025safree,zhang2026differential}.
These methods may improve safety in a hosted setting, but they are fragile in the open-weight setting because an attacker can simply remove the external module and run the base generator directly~\cite{li2024safegen}.
The second category edits the model parameters so that the safety behavior remains after release~\cite{gandikota2023erasing,kumari2023ablating,gandikota2024unified,lu2024mace,gong2024reliable,lyu2024one,li2024safegen,zhang2024defensive,chen2025trce}.
This direction is more durable, but current weight-editing methods still leave an important gap.

The key issue is the training--inference inconsistency.
A text-to-image generator is ultimately deployed as a sampler: starting from noise, it repeatedly applies a learned reverse-time update until an image is produced~\cite{song2020score,Rombach_2022_CVPR}.
However, existing editing methods are typically supervised on states that are not the states visited by the edited model itself during sampling.
Some methods rely on fixed synthetic states generated by the pretrained model; others apply closed-form edits or optimize local objectives without matching the edited model's own rollout trajectory.
As a result, the edited sampler can remain misaligned with the intended safe behavior on the states it actually encounters at inference time, leaving concept erasure incomplete.

In this paper, we formulate concept erasure at the level of the reverse process actually executed after release.
We view text-to-image generation as a unified reverse process, shared across diffusion~\cite{ho2020denoising,Rombach_2022_CVPR} and flow-matching~\cite{lipman2022flow,liu2022flow} model families, and define erasure as follows: on the states visited during sampling, the model under an erase condition should execute the same one-step reverse update as under a safe target condition.
This perspective shifts the problem from static parameter modification to trajectory-level alignment of the deployed sampler.

Based on this formulation, we propose \textbf{OPSD-Erase}, an \emph{on-policy self-distillation} framework for concept erasure.
At training time, the current student model is rolled out from noise, with supervision constructed on student-visited states.
For erase prompts, the student's one-step reverse update is matched to a frozen reference model under a safe target prompt; for anchor prompts that do not request the erased concept, the student is matched to the same frozen reference under the original prompt, preserving benign generation behavior.
This on-policy design directly addresses the training--inference inconsistency that affects prior editing methods.

To make this mechanism explicit, we introduce a trajectory-level probe that measures, along the sampling trajectory, how far the edited sampler departs from the pretrained trajectory and how much one-step reverse-update mismatch remains relative to the safe target.
The probe reveals two characteristic failure patterns in existing methods.
Under fixed-data off-policy supervision, update error grows as the rollout moves beyond the training support.
More generally, non-on-policy editors retain substantial residual mismatch on the states they visit.
In contrast, OPSD-Erase keeps the one-step reverse update aligned on student-visited states.

We evaluate OPSD-Erase on both SD1.4~\cite{Rombach_2022_CVPR} and Z-Image~\cite{team2025zimage}, covering diffusion and flow-matching generators.
At the safety level, OPSD-Erase more reliably suppresses unsafe concept realization under both user prompts and adversarial prompts while preserving benign utility.
At the robustness level, the safety behavior remains stable under wording shift, multi-concept erasure, and sequential extension of the erase list.
Together, these results suggest that effective concept erasure in open-weight text-to-image models depends critically on aligning the sampler's one-step reverse update on the sampling trajectory it actually executes.

Our contributions are as follows:
\begin{itemize}
    \item We formulate concept erasure for text-to-image generation as alignment of the \emph{one-step reverse update on the states visited during sampling}, providing a trajectory-level view of safety for open-weight generators~(Sec.~\ref{sec:problem}).
    \item We propose \textbf{OPSD-Erase}, an on-policy self-distillation framework that trains on student-visited states, uses a frozen reference model to provide safe one-step targets, and preserves benign behavior through anchor conditions~(Sec.~\ref{sec:method}).
    \item We introduce a trajectory-level probe that exposes the training--inference inconsistency of existing editors, and show that OPSD-Erase substantially reduces this mismatch on the sampling trajectory~(Sec.~\ref{sec:mechanism}).
    \item We demonstrate on SD1.4 and Z-Image that OPSD-Erase suppresses unsafe concept realization while preserving benign utility, and remains effective under wording shift, multi-concept erasure, and sequential erasure~(Secs.~\ref{sec:safety}--\ref{sec:generality}).
\end{itemize}

\section{Related Work}
\label{sec:related}

\subsection{Text-to-Image Generative Models}

Modern text-to-image generation is dominated by diffusion models~\cite{saharia2022photorealistic,ICLR2024_081b0806} and, more recently, flow-matching models~\cite{esser2024scaling,wu2025qwen}.
A diffusion model learns a reverse chain that starts from Gaussian noise and denoises it step by step~\cite{ho2020denoising}; latent diffusion runs that chain in an autoencoder latent space and conditions it on text~\cite{Rombach_2022_CVPR}.
Flow matching regresses a velocity on paths from noise to data, and the paths used by diffusion models are instances of this family~\cite{lipman2022flow,albergo2025stochastic}.
Z-Image is a text-conditioned generator trained with that velocity objective~\cite{team2025zimage}.
Although these model families differ in parameterization and solver design, they share the same deployment abstraction: starting from noise, a sampler repeatedly applies learned reverse-time updates until an image is produced.
Our work builds on this view and formulates concept erasure at the level of the reverse process actually executed after release.

\subsection{Concept Erasure for Text-to-Image Models}

Existing defenses for unsafe text-to-image generation broadly follow two routes.
One route adds safety outside the released model.
Prompt-side filters score the request or its text embeddings before sampling~\cite{markov2023holistic,yang2024guardt2i,liu2024latent}.
Output-side checkers score the finished image, including the safety filter shipped with Stable Diffusion and classifiers of inappropriate visual content~\cite{rando2022red,schramowski2022can}.
Inference-time guidance adds a safety term during sampling without changing model weights, as shown by methods such as Safe Latent Diffusion and SAFREE~\cite{schramowski2023safe,yoon2025safree}.
These methods can reduce unsafe outputs in hosted settings, but they rely on the continued presence of the safety pipeline.
Once the code and weights are public, that pipeline can be removed and the base sampler run directly~\cite{gandikota2023erasing,li2024safegen}.

The other route edits the model parameters themselves so that the released generator is less able to realize the target concept.
Fine-tuning-based erasure updates the generative network, most often its attention layers or a small adapter~\cite{gandikota2023erasing,kumari2023ablating,lyu2024one,li2024safegen,chen2025trce}.
AdvUnlearn is the variant that instead adversarially fine-tunes the text encoder~\cite{zhang2024defensive}.
Closed-form editing rewrites attention parameters directly, and can be iterated or combined with a low-rank update when many concepts are removed together~\cite{gandikota2024unified,gong2024reliable,lu2024mace}.
The edited concepts cover unsafe content, artistic styles, identities, and object classes.
Our work belongs to this second line, but differs in focusing on the deployed sampler's behavior on the states it actually visits during sampling.

\subsection{On-Policy Post-Training for Generative Models}

A recurring issue in sequential prediction is the mismatch between the states used for training and the states induced by the learned system at test time. In imitation learning, training only on expert states can lead to compounding error once the learner departs from that support, while Dataset Aggregation reduces this mismatch by rolling out the learner and labeling the states it actually reaches~\cite{ross2010efficient,ross2011reduction}. On-policy reinforcement learning follows the same principle: methods such as proximal policy optimization repeatedly sample trajectories from the current policy and update the policy on those sampled states~\cite{schulman2017proximal}. Similar ideas have recently appeared in distillation, where supervision is provided on sequences generated by the student itself rather than on a fixed external dataset~\cite{agarwal2024policy,zhao2026self}.

Related ideas have also emerged in post-training for generative models. Recent on-policy distillation methods for diffusion and flow-matching models evaluate a teacher on latent states visited during the current student rollout, instead of relying on a fixed set of noised training states~\cite{li2026diffusionopd,jiang2026d}. Online reinforcement learning for these generators likewise collects samples from the current model: Flow-GRPO applies policy-gradient optimization to trajectories drawn from the current flow-based generator~\cite{liu2026flow}, while DiffusionNFT improves reward by optimizing on samples produced by the current model~\cite{zheng2026diffusionnft}. Other post-training methods, such as consistency distillation, also optimize generation behavior after pretraining, though mainly for acceleration rather than safety editing~\cite{song2023consistency}.
Our work is closest in spirit to on-policy post-training, but differs in that it distills the student’s reverse updates across conditions on student-visited states, takes a frozen pretrained generator under a safe target condition as reference, and jointly employs anchor distillation to preserve benign generation utility.

\section{Preliminaries}
\label{sec:preliminaries}

Our method is built on a simple observation: once an open-weight text-to-image model is released, the deployed object is the sampler itself.
A user provides a condition \(c\), the model starts from noise, and generation proceeds by repeatedly applying a learned reverse-time update.
Concept erasure must therefore be formulated at the level of the one-step reverse update executed during sampling.
This section introduces the shared reverse process, the continuous-time dynamics underlying it, and the solvers that instantiate it.

\subsection{A Unified Reverse Process}
\label{subsec:prelim_unified_reverse}

Despite differences in parameterization, modern text-to-image generators share the same sampling abstraction~\cite{song2020score,ho2020denoising,Rombach_2022_CVPR,lipman2022flow,liu2022flow}. They generate an image by iterating a reverse-time transition on a decreasing time grid
\begin{equation}
1 = t_0 > t_1 > \cdots > t_N = 0,
\label{eq:time_grid}
\end{equation}
where \(t=1\) denotes pure noise and \(t=0\) denotes clean data.
Sampling starts from
\begin{equation}
x_{t_0} \sim \mathcal{N}(0,I),
\label{eq:init_noise}
\end{equation}
and each reverse step updates the current latent toward lower noise.
At step \(j\), conditioned on prompt \(c\), the model induces
\begin{equation}
x_{t_{j+1}} \sim p_\theta(\,\cdot \mid x_{t_j}, t_j, t_{j+1}, c),
\qquad j=0,\dots,N-1.
\label{eq:unified_reverse_transition}
\end{equation}
If the transition is deterministic, Eq.~\eqref{eq:unified_reverse_transition} becomes
\begin{equation}
x_{t_{j+1}}
=
\mu_\theta(x_{t_j}, t_j, t_{j+1}, c).
\label{eq:unified_sampling_update}
\end{equation}

Here, \(\theta\) denotes the model parameters, \(p_\theta\) denotes the one-step reverse transition kernel, and \(\mu_\theta\) denotes the corresponding deterministic reverse update.

This reverse process is the common abstraction for generation across model families.
Different model families and different samplers instantiate different forms of \(p\) or \(\mu\), but the deployment behavior always has the same structure, with the model repeatedly applying a one-step reverse update to the states visited by its own rollout.
This shared view is what allows us to define concept erasure before committing to a particular parameterization or solver.

\subsection{Continuous-Time Dynamics}
\label{subsec:prelim_dynamics}

The reverse transition in Eq.~\eqref{eq:unified_reverse_transition} is the discrete object executed at deployment, but it is commonly derived from an underlying continuous-time dynamics. Two widely used formulations are stochastic differential equations (SDEs)~\cite{song2020score} and ordinary differential equations (ODEs)~\cite{lipman2022flow}.

In the stochastic formulation, reverse generation is described by
\begin{equation}
d x_t = f_\theta(x_t,t,c)\,dt + g(t)\,d w_t,
\label{eq:general_sde}
\end{equation}
where \(f_\theta\) is the drift, \(g(t)\) controls the noise scale, and \(w_t\) is a Wiener process. Given the current state, the next state remains random, so each reverse step in Eq.~\eqref{eq:unified_reverse_transition} is a non-degenerate conditional distribution. Diffusion models are naturally associated with this stochastic formulation. A standard forward perturbation path is
\begin{equation}
x_t = \alpha_t x_0 + \sigma_t \epsilon,
\qquad
x_0 \sim p_{\mathrm{data}}(x),\quad
\epsilon \sim \mathcal{N}(0,I),
\label{eq:diff_forward_process}
\end{equation}
which implies
\begin{equation}
q(x_t \mid x_0)
=
\mathcal{N}\!\left(x_t; \alpha_t x_0, \sigma_t^2 I\right).
\label{eq:diff_forward_conditional}
\end{equation}
A common parameterization predicts the injected noise via \(\hat{\epsilon}_\theta(x_t,t,c)\)~\cite{ho2020denoising,Rombach_2022_CVPR}, trained by
\begin{equation}
\mathcal{L}_{\epsilon}
=
\mathbb{E}_{x_0,\epsilon,t,c}
\left[
\left\|
\hat{\epsilon}_\theta(x_t,t,c)-\epsilon
\right\|_2^2
\right].
\label{eq:eps_loss}
\end{equation}

In the deterministic formulation, reverse generation is described by
\begin{equation}
\frac{d x_t}{dt} = v_\theta(x_t,t,c),
\label{eq:general_ode}
\end{equation}
where \(v_\theta\) is a learned velocity field. Given the current state, the infinitesimal evolution is deterministic, and the reverse transition reduces to the update in Eq.~\eqref{eq:unified_sampling_update}. Flow matching is the representative family we consider in this setting. Using the same path parameterization,
\begin{equation}
x_t = \alpha_t x_0 + \sigma_t \epsilon,
\qquad
x_0 \sim p_{\mathrm{data}}(x),\quad
\epsilon \sim \mathcal{N}(0,I),
\label{eq:fm_forward_process}
\end{equation}
with boundary conditions
\begin{equation}
\alpha_0 = 1,\quad \sigma_0 = 0,
\qquad
\alpha_1 = 0,\quad \sigma_1 = 1,
\label{eq:fm_boundary}
\end{equation}
the target velocity is
\begin{equation}
v^\star(x_t,t;x_0,\epsilon)
=
\dot{\alpha}_t x_0 + \dot{\sigma}_t \epsilon.
\label{eq:fm_target_velocity}
\end{equation}
The model is trained by
\begin{equation}
\mathcal{L}_{v}
=
\mathbb{E}_{x_0,\epsilon,t,c}
\left[
\left\|
v_\theta(x_t,t,c)-v^\star(x_t,t;x_0,\epsilon)
\right\|_2^2
\right].
\label{eq:fm_v_loss}
\end{equation}

These two families therefore differ in their continuous-time dynamics, but both induce the same discrete reverse process considered above.

\subsection{Solver Instantiations of the Reverse Update}
\label{subsec:prelim_solver}

Practical generation requires discretizing the continuous-time dynamics on the finite time grid in Eq.~\eqref{eq:time_grid}. Different samplers therefore induce different forms of the one-step reverse update \(\mu\).

For diffusion models with \(\epsilon\)-prediction, a common deterministic discretization is the DDIM update~\cite{song2020denoising}:
\begin{equation}
\begin{aligned}
\mu_\theta^{(\epsilon)}(x_{t_j}, t_j, t_{j+1}, c)
&=
\alpha_{t_{j+1}}
\frac{x_{t_j}-\sigma_{t_j}\hat{\epsilon}_\theta(x_{t_j},t_j,c)}{\alpha_{t_j}}
\\
&\quad+
\sigma_{t_{j+1}}\hat{\epsilon}_\theta(x_{t_j},t_j,c).
\end{aligned}
\label{eq:eps_update}
\end{equation}
Other diffusion samplers, such as DDPM-style stochastic updates~\cite{ho2020denoising}, instantiate the same reverse transition with different step rules.

For flow matching models, which directly predict the velocity field, a common discretization is the Euler update~\cite{lipman2022flow}:
\begin{equation}
\mu_\theta^{(v)}(x_{t_j}, t_j, t_{j+1}, c)
=
x_{t_j} + v_\theta(x_{t_j}, t_j, c)\,(t_{j+1}-t_j).
\label{eq:v_update}
\end{equation}
Higher-order ODE solvers instantiate the same reverse transition with different discretization rules.

Thus, diffusion and flow-matching models differ in their continuous-time dynamics and solver forms, but share a unified reverse process on the same discrete grid.

\section{Problem Formulation and Threat Model}
\label{sec:problem}

\subsection{Problem Formulation}
\label{subsec:problem_formulation}

The preliminaries identify the reverse transition as the common object across model families and solvers.
We now formulate concept erasure as an alignment problem on that one-step reverse update at the states visited during inference.

Let \(c_1\) denote the erase condition, namely the prompt associated with the unsafe concept to be erased.
Let \(c_2\) denote the safe target, namely the prompt specifying the desired post-erasure behavior.
For a given state \(x_{t_j}\) on the reverse-time grid and the next time \(t_{j+1}\), the model induces a one-step reverse transition kernel
\begin{equation}
p_\theta(\,\cdot \mid x_{t_j}, t_j, t_{j+1}, c).
\label{eq:ce_transition_kernel}
\end{equation}

The ideal erasure requirement is that, at the same current state \(x_{t_j}\), the model under the erase condition \(c_1\) should induce the same reverse transition as under the safe target \(c_2\):
\begin{equation}
p_\theta(\,\cdot \mid x_{t_j}, t_j, t_{j+1}, c_1)
=
p_\theta(\,\cdot \mid x_{t_j}, t_j, t_{j+1}, c_2).
\label{eq:ce_self_alignment}
\end{equation}

In practice, the post-erasure target behavior is not directly available before training.
We therefore introduce a frozen reference model \(p_{\mathrm{ref}}\) and use its safe transition under \(c_2\) as the target:
\begin{equation}
p_\theta(\,\cdot \mid x_{t_j}, t_j, t_{j+1}, c_1)
\approx
p_{\mathrm{ref}}(\,\cdot \mid x_{t_j}, t_j, t_{j+1}, c_2).
\label{eq:ce_ref_alignment}
\end{equation}

In the deterministic case introduced in Sec.~\ref{subsec:prelim_unified_reverse}, the transition kernel reduces to a one-step reverse update, and Eq.~\eqref{eq:ce_ref_alignment} becomes
\begin{equation}
\mu_\theta(x_{t_j}, t_j, t_{j+1}, c_1)
\approx
\mu_{\mathrm{ref}}(x_{t_j}, t_j, t_{j+1}, c_2).
\label{eq:ce_mu_alignment}
\end{equation}
Thus, in both stochastic and deterministic settings, concept erasure is defined as aligning the one-step reverse update under \(c_1\) with the safe behavior under \(c_2\) on the states visited during sampling.

This state dependence is essential.
The released model does not execute its reverse update on arbitrary training states, but on the states visited by its own rollout.
If the states used to construct the defense differ from the states visited during inference, the resulting sampler may remain misaligned with the safe target on its deployed trajectory.
This is the training--inference inconsistency at the core of the problem.

\subsection{Threat Model}
\label{subsec:threat_model}

We consider the open-weight deployment setting for text-to-image generation.
The defender may edit the model parameters before release, but once the weights are public, an attacker can deploy and query the model independently.
This setting rules out defenses that rely on safety components remaining in the serving pipeline.
In particular, external detection modules and inference-time guidance can be omitted by an attacker who runs the released model on their own system.

The attacker interacts with the model through text prompts and aims to induce unsafe concepts during generation.
These may include, but are not limited to, sexual or other NSFW content, graphic violence, self-harm, hateful or harassing content, and illegal or otherwise harmful activities~\cite{schramowski2023safe,yang2024mma,chin2023prompting4debugging,tsai2024ring,zhang2024generate}.
The attacker may vary the prompting pattern through paraphrases, contextual composition, or out-of-distribution wording, but does not modify the released model parameters.
In particular, we exclude post-release fine-tuning from our threat model, because once an attacker is allowed to retrain the released model, erased or filtered concepts can in principle be reintroduced~\cite{crectu2026evaluating}.

Under this threat model, the defender's goal is to edit the model so that unsafe concepts are no longer reliably realized during sampling, while benign generation behavior remains intact.
A successful defense should therefore satisfy two requirements.
First, prompts for the erase condition and its variants should no longer reliably induce the unsafe concept.
Second, prompts that do not request the erased concept should preserve benign utility as much as possible.

\section{Method}
\label{sec:method}

\begin{figure*}[t]
    \centering
    \includegraphics[width=0.88\linewidth]{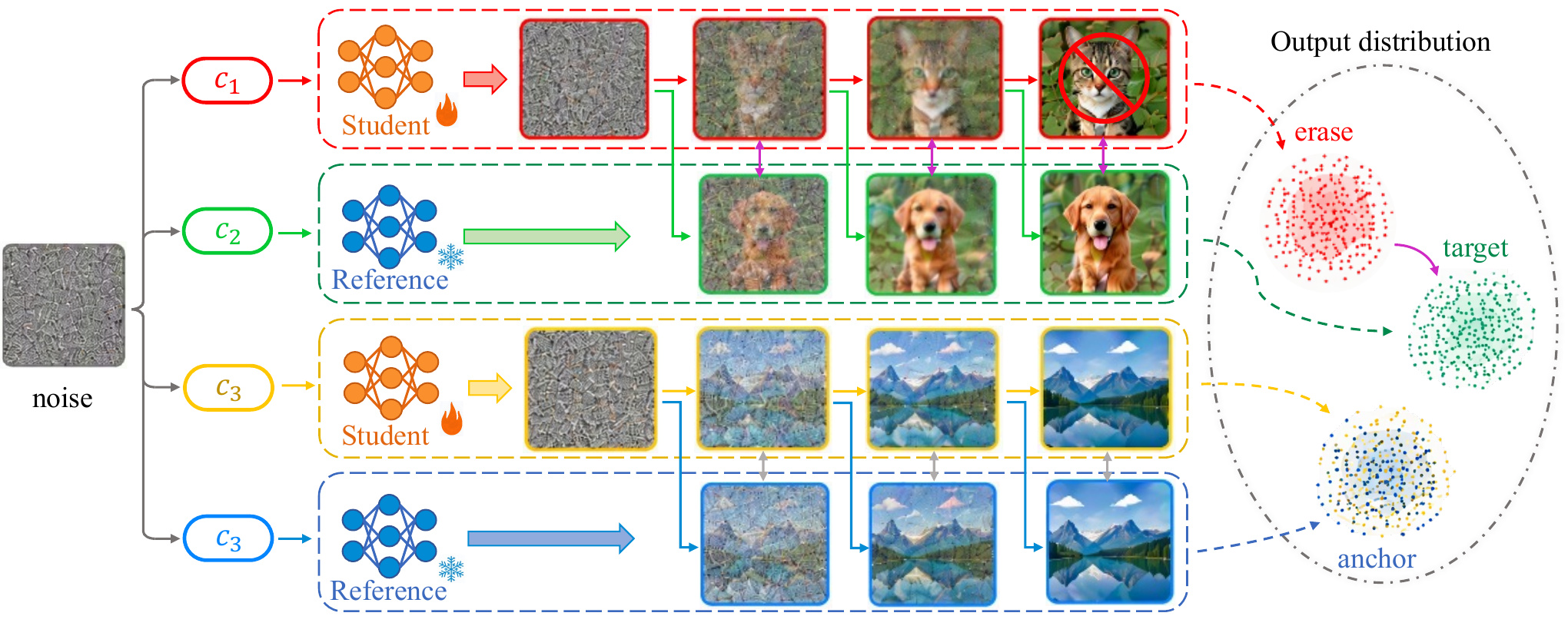}
    \caption{Overview of OPSD-Erase.
    From a noise sample, the student is rolled out under either an erase condition \(c_1\) or an anchor condition \(c_3\).
    For erase samples, the student's one-step reverse update under \(c_1\) is matched to the frozen reference under the safe target \(c_2\).
    For anchor samples, the student's update under \(c_3\) is matched to the frozen reference under the same \(c_3\), preserving benign generation behavior.
    All supervision is imposed on student-visited states.
    Conceptually, this pulls the erase output distribution toward the safe target distribution while keeping the anchor distribution unchanged.}
    \label{fig:method}
\end{figure*}

Building on Secs.~\ref{subsec:problem_formulation} and \ref{subsec:threat_model}, we seek a durable defense that remains effective after the model weights are released.
Our key design choice is to train on student-visited states, rather than on a fixed external state distribution.
At each iteration, the current student is rolled out from noise, and a frozen reference model provides one-step supervision on the resulting states.
For erase samples, the student under the erase condition \(c_1\) is aligned to the reference under the safe target \(c_2\).
For anchor samples, the student under an anchor condition \(c_3\) is aligned to the reference under the same \(c_3\) to preserve benign generation behavior.
Because the student is supervised by a frozen reference from the same pretrained model on student-visited states, we refer to this procedure as \textbf{on-policy self-distillation}.

We first derive the objective in the stochastic setting, where one-step reverse-transition alignment can be written as a per-step Kullback--Leibler divergence.
We then specialize it to deterministic samplers.
Finally, we instantiate the resulting loss under the reverse update solvers introduced in Sec.~\ref{subsec:prelim_solver}.

\subsection{Stochastic Reverse-Transition Alignment}
\label{subsec:method_sde}

We begin with the stochastic case.
A one-step reverse transition can be written as
\begin{equation}
x_{t_{j+1}}
=
\mu(x_{t_j}, t_j, t_{j+1}, c)
+
\bar{\sigma}_j \varepsilon_j,
\qquad
\varepsilon_j \sim \mathcal{N}(0,I),
\label{eq:method_sde_step}
\end{equation}
where \(\bar{\sigma}_j \ge 0\) is the step noise induced by the sampler.
This gives the Gaussian reverse kernel
\begin{equation}
p(\,\cdot \mid x_{t_j}, t_j, t_{j+1}, c)
=
\mathcal{N}\!\left(
\mu(x_{t_j}, t_j, t_{j+1}, c),
\bar{\sigma}_j^2 I
\right).
\label{eq:method_gaussian_kernel}
\end{equation}

Under the formulation in Eq.~\eqref{eq:ce_ref_alignment}, the student under the erase condition $c_1$ should match the frozen reference under the safe target $c_2$ at the same visited state $x_{t_j}$.
We therefore write
\begin{equation}
p_\theta(\,\cdot \mid x_{t_j}, t_j, t_{j+1}, c_1)
=
\mathcal{N}\!\left(
\mu_\theta(x_{t_j}, t_j, t_{j+1}, c_1),
\bar{\sigma}_j^2 I
\right),
\label{eq:method_student_kernel}
\end{equation}
and
\begin{equation}
p_{\mathrm{ref}}(\,\cdot \mid x_{t_j}, t_j, t_{j+1}, c_2)
=
\mathcal{N}\!\left(
\mu_{\mathrm{ref}}(x_{t_j}, t_j, t_{j+1}, c_2),
\bar{\sigma}_j^2 I
\right).
\label{eq:method_ref_kernel}
\end{equation}
Because the two kernels share the same covariance, their discrepancy is entirely determined by the gap between their means.

For two Gaussians with the same covariance $\Sigma$,
\begin{equation}
\mathrm{KL}\!\left(
\mathcal{N}(\mu_1,\Sigma)
\,\|\,
\mathcal{N}(\mu_2,\Sigma)
\right)
=
\frac{1}{2}
(\mu_1-\mu_2)^\top
\Sigma^{-1}
(\mu_1-\mu_2).
\label{eq:method_same_cov_kl}
\end{equation}
Substituting $\Sigma=\bar{\sigma}_j^2 I$ yields
\begin{align}
&\mathrm{KL}\!\Bigl(
p_\theta(\cdot \mid x_{t_j}, t_j, t_{j+1}, c_1)
\,\|\,
p_{\mathrm{ref}}(\cdot \mid x_{t_j}, t_j, t_{j+1}, c_2)
\Bigr)
\notag\\
&=
\frac{
\bigl\|
\mu_\theta(x_{t_j}, t_j, t_{j+1}, c_1)
-
\mu_{\mathrm{ref}}(x_{t_j}, t_j, t_{j+1}, c_2)
\bigr\|_2^2
}{
2\bar{\sigma}_j^2
}.
\raisetag{10pt}
\label{eq:method_perstep_kl}
\end{align}

Following DiffusionOPD~\cite{li2026diffusionopd}, the shared-covariance KL reduces to this mean matching; relative to it, our reference is a frozen copy of the pretrained model under a different condition, yielding on-policy self-distillation. The key point is that this discrepancy is evaluated on student-visited states.
Let
\begin{equation}
x_{t_0:t_N} \sim p_\theta(\cdot \mid c_1)
\label{eq:method_student_rollout}
\end{equation}
denote a student rollout under the erase condition $c_1$.
The resulting on-policy stochastic erasure objective is
\begin{align}
\mathcal{L}_{\mathrm{erase}}^{\mathrm{SDE}}(\theta)
&=
\mathbb{E}_{x_{t_0:t_N}\sim p_\theta(\cdot \mid c_1)}
\Biggl[
\sum_{j=0}^{N-1}
\frac{1}{2\bar{\sigma}_j^2}
\Bigl\|
\mu_\theta(x_{t_j}, t_j, t_{j+1}, c_1)
\notag\\
&\qquad\qquad
-
\mu_{\mathrm{ref}}(x_{t_j}, t_j, t_{j+1}, c_2)
\Bigr\|_2^2
\Biggr].
\raisetag{10pt}
\label{eq:method_sde_loss}
\end{align}
This is the expected one-step reverse-transition KL between the student under \(c_1\) and the safe reference under \(c_2\), taken over student-visited states.

\subsection{Deterministic Specialization}
\label{subsec:method_ode}

In practice, deterministic samplers are used by most open-weight text-to-image models.
When the step noise is removed, Eq.~\eqref{eq:method_sde_step} reduces to
\begin{equation}
x_{t_{j+1}}
=
\mu(x_{t_j}, t_j, t_{j+1}, c).
\label{eq:method_det_step}
\end{equation}
In this limit, the covariance in Eq.~\eqref{eq:method_perstep_kl} becomes singular, so the variance-scaled KL expression is no longer directly usable.
What remains well-defined is the mismatch between the one-step reverse updates themselves.
We therefore retain the same mean-matching structure and define the deterministic erasure loss as
\begin{align}
\mathcal{L}_{\mathrm{erase}}^{\mathrm{ODE}}(\theta)
&=
\mathbb{E}_{x_{t_0:t_N}\sim p_\theta^{\mathrm{ODE}}(\cdot \mid c_1)}
\Biggl[
\sum_{j=0}^{N-1}
\frac{1}{2}
\Bigl\|
\mu_\theta(x_{t_j}, t_j, t_{j+1}, c_1)
\notag\\
&\qquad\qquad
-
\mu_{\mathrm{ref}}(x_{t_j}, t_j, t_{j+1}, c_2)
\Bigr\|_2^2
\Biggr].
\raisetag{10pt}
\label{eq:method_ode_loss}
\end{align}
Equation~\eqref{eq:method_ode_loss} is the deterministic form of Eq.~\eqref{eq:ce_mu_alignment} and serves as the core erasure objective in our method.

\subsection{Anchor-Based Preservation}
\label{subsec:method_preservation}

The erasure loss alone does not constrain prompts that do not request the erased concept.
To preserve benign generation behavior, we introduce an anchor condition \(c_3\), where \(c_3\) does not request the erased concept.
For such anchor samples, the student is required to remain close to the frozen reference under the same condition $c_3$.
Using the same deterministic one-step matching principle, we define
\begin{align}
\mathcal{L}_{\mathrm{preserve}}^{\mathrm{ODE}}(\theta)
&=
\mathbb{E}_{x_{t_0:t_N}\sim p_\theta^{\mathrm{ODE}}(\cdot \mid c_3)}
\Biggl[
\sum_{j=0}^{N-1}
\frac{1}{2}
\Bigl\|
\mu_\theta(x_{t_j}, t_j, t_{j+1}, c_3)
\notag\\
&\qquad\qquad
-
\mu_{\mathrm{ref}}(x_{t_j}, t_j, t_{j+1}, c_3)
\Bigr\|_2^2
\Biggr].
\raisetag{10pt}
\label{eq:method_preserve_loss}
\end{align}

Conceptually, the full objective can be viewed as
\begin{equation}
\mathcal{L}_{\mathrm{total}}(\theta)
=
\mathcal{L}_{\mathrm{erase}}^{\mathrm{ODE}}(\theta)
+
\lambda_{\mathrm{pres}}
\mathcal{L}_{\mathrm{preserve}}^{\mathrm{ODE}}(\theta),
\label{eq:method_total_loss}
\end{equation}
where $\lambda_{\mathrm{pres}} \ge 0$ controls the relative strength of preservation.

In practice, we do not explicitly evaluate the two expectations in Eq.~\eqref{eq:method_total_loss} within the same minibatch and then apply a scalar weight.
Instead, we mix erase pairs $(c_1,c_2)$ with anchor pairs $(c_3,c_3)$ during training.
The sampling frequency and batch share of anchor samples determine the effective preservation strength.

\subsection{Solver Instantiations}
\label{subsec:method_instantiation}

The outer training objective in Eq.~\eqref{eq:method_total_loss} is shared across model families.
What changes is only the form of the one-step reverse update $\mu$.

For diffusion models with $\epsilon$-prediction, we first form the clean-sample estimate
\begin{equation}
\hat{x}_{0,\theta}(x_{t_j}, t_j, c)
=
\frac{x_{t_j}-\sigma_{t_j}\hat{\epsilon}_\theta(x_{t_j}, t_j, c)}{\alpha_{t_j}}.
\label{eq:method_x0_est}
\end{equation}
The DDIM update~\cite{song2020denoising} is then
\begin{equation}
\begin{aligned}
\mu_\theta^{(\epsilon)}(x_{t_j}, t_j, t_{j+1}, c)
&=
\alpha_{t_{j+1}}\hat{x}_{0,\theta}(x_{t_j}, t_j, c)
\\
&\quad+
\sigma_{t_{j+1}}\hat{\epsilon}_\theta(x_{t_j}, t_j, c).
\end{aligned}
\label{eq:method_eps_inst}
\end{equation}
Replacing $\hat{\epsilon}_\theta$ with $\hat{\epsilon}_{\mathrm{ref}}$ gives the corresponding reference update $\mu_{\mathrm{ref}}^{(\epsilon)}$.
The per-step matching loss is
\begin{align}
\ell_j^{(\epsilon)}(c_a,c_b)
&=
\frac{1}{2}
\Bigl\|
\mu_\theta^{(\epsilon)}(x_{t_j}, t_j, t_{j+1}, c_a)
\notag\\
&\qquad\qquad
-
\mu_{\mathrm{ref}}^{(\epsilon)}(x_{t_j}, t_j, t_{j+1}, c_b)
\Bigr\|_2^2.
\label{eq:method_eps_loss}
\end{align}

For flow-matching models, let $\Delta t_j = t_{j+1}-t_j$.
The Euler update~\cite{lipman2022flow} is
\begin{equation}
\mu_\theta^{(v)}(x_{t_j}, t_j, t_{j+1}, c)
=
x_{t_j} + \Delta t_j\, v_\theta(x_{t_j}, t_j, c).
\label{eq:method_v_inst}
\end{equation}
Replacing $v_\theta$ with $v_{\mathrm{ref}}$ gives the reference update $\mu_{\mathrm{ref}}^{(v)}$.
The per-step matching loss is
\begin{align}
\ell_j^{(v)}(c_a,c_b)
&=
\frac{1}{2}
\Bigl\|
\mu_\theta^{(v)}(x_{t_j}, t_j, t_{j+1}, c_a)
\notag\\
&\qquad\qquad
-
\mu_{\mathrm{ref}}^{(v)}(x_{t_j}, t_j, t_{j+1}, c_b)
\Bigr\|_2^2.
\label{eq:method_v_loss}
\end{align}
In both cases, $(c_a,c_b)=(c_1,c_2)$ gives the erasure loss, while $(c_a,c_b)=(c_3,c_3)$ gives the preservation loss.

\subsection{On-Policy Self-Distillation}
\label{subsec:method_opsd}

Algorithm~\ref{alg:opsd} summarizes the full training procedure.
During training, we sample erase prompts \(c_1\) together with their corresponding safe target prompts \(c_2\), as well as anchor prompts \(c_3\) that do not request the erased concept.
The current student is then rolled out from noise on a mixed batch containing erase pairs \((c_1,c_2)\) and anchor pairs \((c_3,c_3)\).

For erase samples, the student is rolled out under \(c_1\) and matched to the reference under \(c_2\); for anchor samples, it is rolled out under \(c_3\) and matched to the reference under the same \(c_3\).
The student is then optimized on the states visited by its own rollout.
Within each optimization stage, these student-visited states are treated as fixed training samples, so the loss supervises the current one-step update at those states rather than backpropagating through the entire rollout.

In our implementation, the student is fine-tuned with LoRA adapters~\cite{hu2021lora} inserted into attention projections, including the query, key, value, and output matrices, together with any additional attention projections present in the architecture.
The reference remains the full frozen pretrained model.
Erase prompts are constructed from simple subject-centric templates so that the erased concept \(c_1\) remains the main visual subject.
The corresponding safe target prompt \(c_2\) is obtained by replacing that subject with the safe target concept while keeping the rest of the template unchanged.
Anchor prompts are sampled from prompts that do not request the erased concept.
Detailed prompt-construction rules are given in Appendix~\ref{app:prompt_construction}. Unless otherwise stated, training uses 10 sampling steps at \(512\times512\) with CFG scale 4~\cite{ho2022classifier}; evaluation uses 50 steps, \(512\times512\), and scale 7.5 for SD1.4, and 40 steps, \(1024\times1024\), and scale 4 for Z-Image.

\begin{algorithm}[t]
\caption{OPSD-Erase}
\label{alg:opsd}
\begin{algorithmic}[1]
\REQUIRE Pretrained model \(f_{\mathrm{pre}}\), erase-condition distribution \(\mathcal{C}_1\), safe-target map \(\mathcal{C}_2(\cdot)\), anchor-condition distribution \(\mathcal{C}_3\), time grid \(\{t_j\}_{j=0}^{N}\)
\ENSURE Student model \(f_\theta\)

\STATE Initialize \(f_\theta \leftarrow f_{\mathrm{pre}}\)
\STATE Initialize frozen reference \(f_{\mathrm{ref}} \leftarrow f_{\mathrm{pre}}\)

\WHILE{not converged}
    \STATE Sample erase conditions \(c_1 \sim \mathcal{C}_1\) and set \(c_2 \leftarrow \mathcal{C}_2(c_1)\)
    \STATE Sample anchor conditions \(c_3 \sim \mathcal{C}_3\)
    \STATE Form a mixed batch of erase pairs \((c_1,c_2)\) and anchor pairs \((c_3,c_3)\)

    \FOR{each pair \((c_a,c_b)\) in the mixed batch}
        \STATE Sample \(x_{t_0} \sim \mathcal{N}(0,I)\)
        \FOR{\(j=0,\dots,N-1\)}
            \STATE \(x_{t_{j+1}} \leftarrow \mu_\theta(x_{t_j}, t_j, t_{j+1}, c_a)\)
        \ENDFOR
        \STATE Store \(x_{t_0},x_{t_1},\dots,x_{t_N}\)
    \ENDFOR

    \STATE Set \(\mathcal{L}\leftarrow 0\)

    \FOR{each stored trajectory with pair \((c_a,c_b)\)}
        \FOR{\(j=0,\dots,N-1\)}
            \STATE \(\mu_{\theta,j} \leftarrow \mu_\theta(x_{t_j}, t_j, t_{j+1}, c_a)\)
            \STATE \(\mu_{\mathrm{ref},j} \leftarrow \mu_{\mathrm{ref}}(x_{t_j}, t_j, t_{j+1}, c_b)\)
            \STATE \(\mathcal{L} \leftarrow \mathcal{L} + \frac12 \|\mu_{\theta,j}-\mu_{\mathrm{ref},j}\|_2^2\)
        \ENDFOR
    \ENDFOR

    \STATE Update \(\theta\) using gradient descent on \(\mathcal{L}\)
\ENDWHILE
\end{algorithmic}
\end{algorithm}

\begin{figure}[t]
    \centering
    \includegraphics[width=\linewidth]{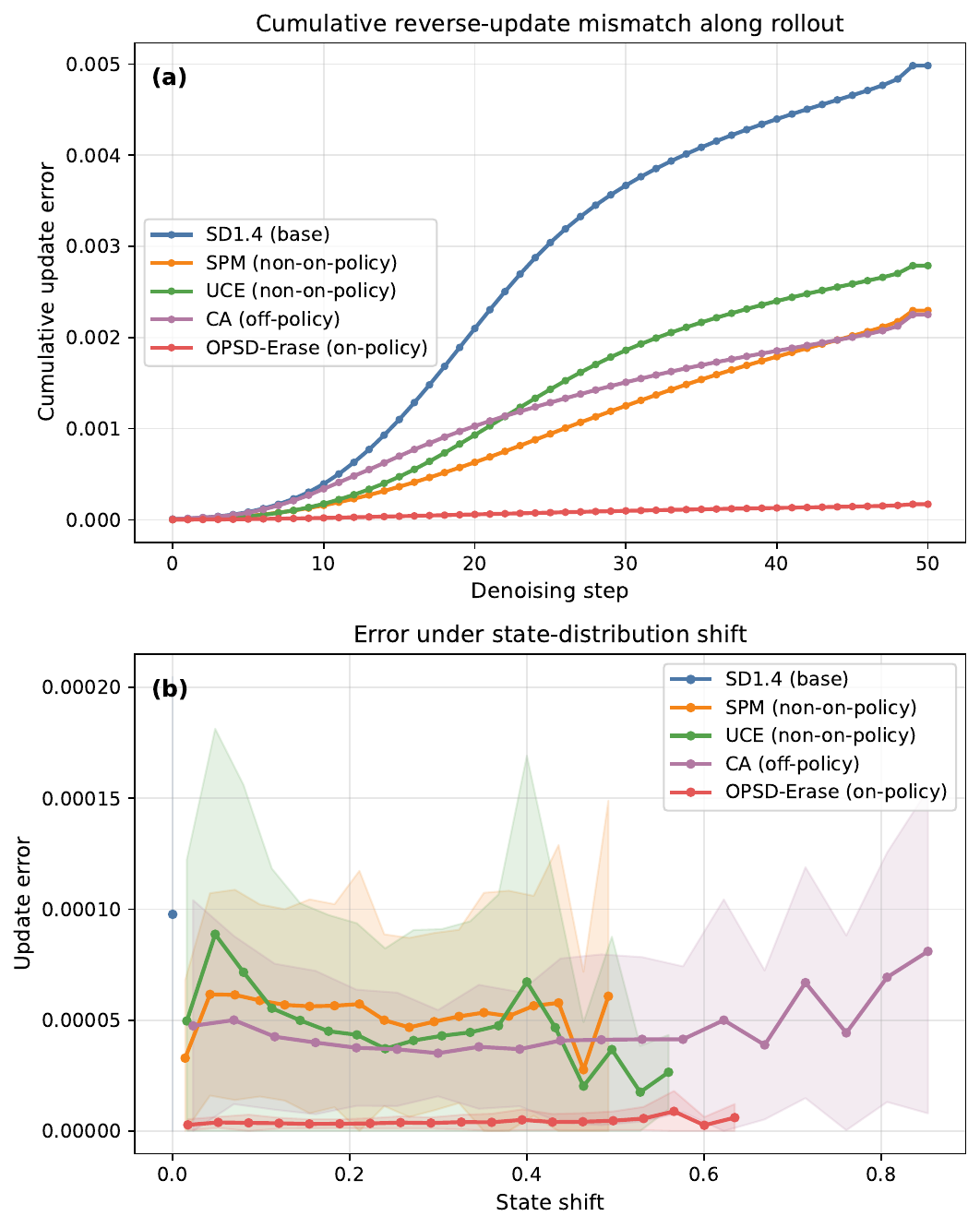}
    \caption{Training--inference inconsistency on SD1.4, measured on the sampling trajectory of the released sampler.
    (a) Cumulative one-step reverse-update mismatch along the rollout.
    All editors reduce mismatch relative to the unedited model, while OPSD-Erase remains consistently low.
    (b) The one-step mismatch plotted against departure from the pretrained rollout.
    Fixed-data off-policy supervision deteriorates once rollout states move beyond its support, whereas non-on-policy editors retain residual mismatch on the states they visit.
    OPSD-Erase stays aligned on student-visited states.}
    \label{fig:mechanism_combined}
\end{figure}

\section{Mechanism-Level Analysis: Training--Inference Inconsistency}
\label{sec:mechanism}

Our starting point is that concept erasure remains incomplete when the edited sampler is not aligned with the safe one-step reverse update on the states it visits during sampling.
Section~\ref{sec:problem} formulated this failure as a mismatch between training and sampling states, and Section~\ref{sec:method} addressed it by training on student-visited states.
We now measure this mechanism directly.

Our analysis is conducted on SD1.4~\cite{Rombach_2022_CVPR} and compares four representative editors: CA~\cite{kumari2023ablating}, UCE~\cite{gandikota2024unified}, SPM~\cite{lyu2024one}, and OPSD-Erase.
CA is a fixed-data off-policy editor, UCE and SPM are non-on-policy editors, and OPSD-Erase is an on-policy editor.
CA is trained on a frozen collection generated from the pretrained model and is therefore tied to a fixed off-policy state distribution.
UCE performs a closed-form parameter edit and does not optimize on any rollout states after editing.
SPM is a fine-tuning-based editor whose supervision is specified independently of the edited sampler’s rollout.
OPSD-Erase instead trains on student-visited states from the current student.

The mechanism-level question is not yet whether an unsafe concept appears in the final image.
It is whether, at test time, the released sampler executes the safe one-step reverse update on the states it reaches.
To answer this, we probe two quantities along the sampling trajectory:
(i) how far the edited rollout has moved away from the pretrained rollout, and
(ii) how far the edited one-step update at that visited state remains from the safe target.
If supervision is imposed on the states the edited sampler actually visits, the one-step mismatch should remain small on those states.
If supervision is tied to a frozen external state distribution, mismatch should grow once the rollout leaves that support.
If the method is a closed-form edit or a fine-tuning-based editor trained without on-policy trajectory matching, residual mismatch should remain on the visited states throughout.

\subsection{Probe Definition}
\label{subsec:mechanism_probe}

For sample \(i\) at reverse step \(j\), let \(x_{i,j}^{(\theta)}\) denote the latent visited by the edited model when rolled out under the erase condition \(c_1\), and let \(x_{i,j}^{(\mathrm{pre})}\) denote the latent produced by the original pretrained model from the same initial noise and prompt.
We use the distance between these two latents to quantify how far inference under the edited sampler has departed from the original rollout:
\begin{equation}
d_{i,j}^{(\theta)}
=
\frac{1}{CHW}
\left\|
x_{i,j}^{(\theta)}-x_{i,j}^{(\mathrm{pre})}
\right\|_2^2,
\label{eq:mechanism_state_shift}
\end{equation}
where \(C,H,W\) are the latent dimensions.

We then evaluate the one-step reverse-update mismatch at that same visited state.
Using the deterministic alignment objective in Eq.~\eqref{eq:ce_mu_alignment}, we define
\begin{align}
e_{i,j}^{(\theta)}
&=
\frac{1}{CHW}
\Bigl\|
\mu_\theta(x_{i,j}^{(\theta)}, t_j, t_{j+1}, c_1)
\notag\\
&\qquad\qquad
-
\mu_{\mathrm{ref}}(x_{i,j}^{(\theta)}, t_j, t_{j+1}, c_2)
\Bigr\|_2^2.
\raisetag{10pt}
\label{eq:mechanism_step_error}
\end{align}
Equation~\eqref{eq:mechanism_step_error} is the test-time counterpart of the erasure discrepancy defined in Section~\ref{sec:problem}: it measures whether, at the state the edited sampler has actually reached, the erase-condition update still differs from the safe target update.

These two quantities expose training--inference inconsistency in complementary ways.
The state shift \(d_{i,j}^{(\theta)}\) measures how far the edited rollout has moved from the pretrained trajectory.
The update error \(e_{i,j}^{(\theta)}\) measures whether the released sampler is still executing the intended safe update at that state.
Different methods can exhibit different mismatch patterns.
Some retain substantial update mismatch even under modest state shift, while others develop sharply increasing mismatch once the rollout moves beyond the support of fixed off-policy supervision.
In either case, incomplete erasure is driven by mismatch in the one-step reverse update on the states visited by the edited sampler.

\begin{table*}[t]
\centering
\caption{Sexual-content erasure on SD1.4.
Some baseline numbers follow the TRCE protocol~\cite{chen2025trce}.}
\label{tab:sex_erasure_sd14}
\small
\setlength{\tabcolsep}{4.5pt}
\renewcommand{\arraystretch}{1.08}
\begin{tabular*}{\textwidth}{@{\extracolsep{\fill}}lccccccc@{}}
\toprule
\multirow{2}{*}{Method} 
& \multicolumn{1}{c}{\textbf{User Prompt}} 
& \multicolumn{4}{c}{\textbf{Adversarial Prompt}} 
& \multicolumn{2}{c}{\textbf{Benign Utility}} \\
\cmidrule(lr){2-2} \cmidrule(lr){3-6} \cmidrule(lr){7-8}
& I2P$\downarrow$ & MMA$\downarrow$ & P4D$\downarrow$ & Ring$\downarrow$ & UnDiff$\downarrow$ & FID$\downarrow$ & CLIP-Score$\uparrow$ \\
\midrule
SD1.4 & 34.69\% & 79.00\% & 83.44\% & 59.49\% & 57.75\% & 27.18 & 30.97 \\
\midrule
SLD & 17.29\% & 64.30\% & 63.58\% & 34.18\% & 47.18\% & 34.28 & 29.78 \\
Safree & 10.20\% & 44.60\% & 49.67\% & 22.38\% & 30.00\% & 30.11 & 26.79 \\
ESD & 31.15\% & 58.50\% & 82.67\% & 50.63\% & 77.46\% & \textbf{26.88} & \textbf{31.21} \\
SPM & 12.57\% & 67.40\% & 63.58\% & 31.65\% & 35.21\% & 27.66 & 30.85 \\
UCE & 8.16\% & 30.80\% & 43.71\% & 13.92\% & 19.72\% & 27.20 & 30.92 \\
SafeGEN & 11.39\% & \underline{0.40\%} & 8.70\% & 13.75\% & 21.98\% & 27.39 & \underline{31.06} \\
RECE & 6.34\% & 23.10\% & 32.00\% & 6.33\% & 15.49\% & 28.26 & 30.79 \\
MACE & 7.09\% & 10.60\% & \underline{7.95\%} & 10.13\% & 11.27\% & 26.98 & 28.84 \\
AdvUnlearn & 1.71\% & \textbf{0.30\%} & \textbf{1.99\%} & 6.33\% & 3.52\% & 29.65 & 28.93 \\
TRCE & \underline{1.29\%} & 1.40\% & \textbf{1.99\%} & \underline{1.27\%} & \underline{0.70\%} & \underline{26.89} & 30.71 \\
\textbf{OPSD-Erase} & \textbf{0.54\%} & 0.60\% & \textbf{1.99\%} & \textbf{0.00\%} & \textbf{0.00\%} & 27.23 & 30.79 \\
\bottomrule
\end{tabular*}
\end{table*}

\begin{table*}[t]
\centering
\caption{Sexual-content erasure on Z-Image.
Benign utility uses the metrics standard for this flow-matching model family~\cite{team2025zimage}.}
\label{tab:sex_erasure_Z-Image}
\small
\setlength{\tabcolsep}{4.2pt}
\renewcommand{\arraystretch}{1.08}
\begin{tabular*}{\textwidth}{@{\extracolsep{\fill}}lcccccccccc@{}}
\toprule
\multirow{2}{*}{Method}
& \multicolumn{1}{c}{\textbf{User Prompt}}
& \multicolumn{4}{c}{\textbf{Adversarial Prompt}}
& \multicolumn{5}{c}{\textbf{Benign Utility}} \\
\cmidrule(lr){2-2} \cmidrule(lr){3-6} \cmidrule(lr){7-11}
& I2P$\downarrow$ & MMA$\downarrow$ & P4D$\downarrow$ & Ring$\downarrow$ & UnDiff$\downarrow$
& DPG$\uparrow$ & GenEval$\uparrow$ & PickScore$\uparrow$ & CLIP-Score$\uparrow$ & HPSv2$\uparrow$ \\
\midrule
Z-Image & 10.63\% & 42.30\% & 46.36\% & 41.77\% & 17.61\% & 86.31 & 61.21 & 82.53 & 28.51 & 28.68 \\
\midrule
ESD-style & 1.83\% & \underline{0.50\%} & \underline{3.97\%} & 19.00\% & 3.52\% & \underline{86.41} & \underline{54.05} & \textbf{81.68} & \underline{27.59} & \textbf{28.58} \\
DVE & \underline{1.50\%} & 0.90\% & 7.28\% & \underline{7.59\%} & \textbf{2.11\%} & 86.35 & 53.58 & 80.73 & \textbf{27.90} & 27.55 \\
\textbf{OPSD-Erase} & \textbf{1.07\%} & \textbf{0.00\%} & \textbf{0.00\%} & \textbf{1.27\%} & \underline{2.81\%} & \textbf{86.45} & \textbf{59.05} & \underline{80.99} & 27.53 & \underline{27.79} \\
\bottomrule
\end{tabular*}
\end{table*}

\begin{figure*}[t]
    \centering
    \includegraphics[width=\textwidth]{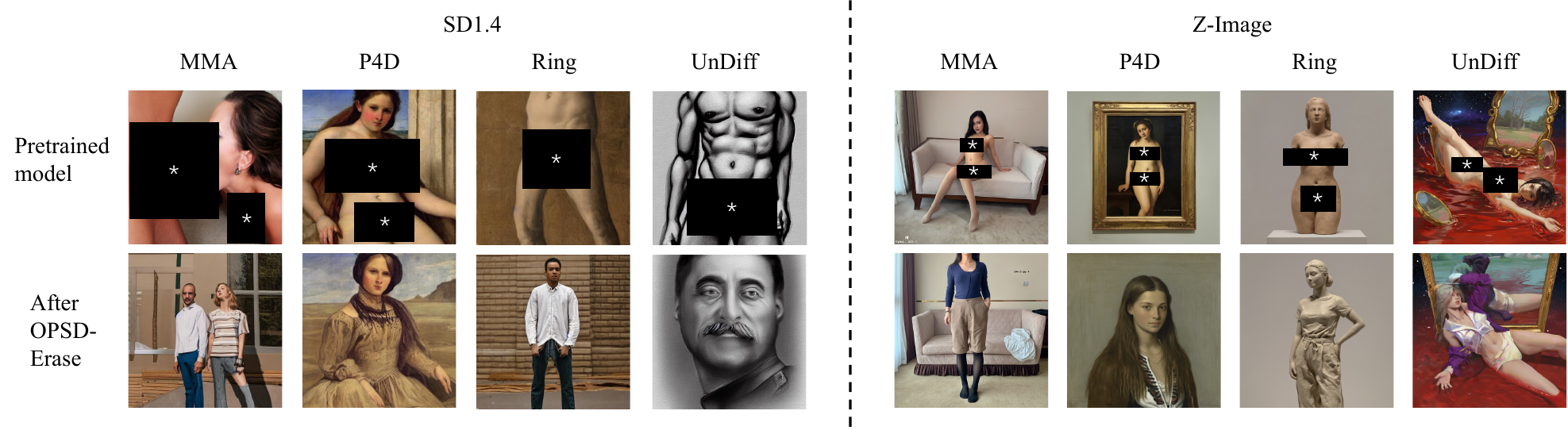}
    \caption{Prompts that request sexual content, before and after OPSD-Erase.
    On both SD1.4 and Z-Image, the unsafe concept is strongly suppressed while the model remains visually coherent.}
    \label{fig:sex_visualization}
\end{figure*}

\subsection{Compared Methods and Setup}
\label{subsec:mechanism_setup}

We instantiate the probe on SD1.4 using the same pretrained checkpoint, prompt set, initial noise, and sampler across all editors.
The compared methods are CA, UCE, SPM, and OPSD-Erase.

For each method, we sample erase prompts \(c_1\), construct the corresponding safe targets \(c_2\), roll out the edited model from shared noise, and record \((d_{i,j}^{(\theta)}, e_{i,j}^{(\theta)})\) at each reverse step.
This gives a trajectory-level view of whether the released sampler remains aligned with the safe one-step reverse update as its own rollout evolves.

\subsection{Main Observations}
\label{subsec:mechanism_results}

Figure~\ref{fig:mechanism_combined} reports the probe on the sampling trajectory.

Panel~(a) accumulates the one-step mismatch in Eq.~\eqref{eq:mechanism_step_error} along the rollout.
This accumulated quantity reflects the mismatch incurred by the sampler during generation.
Relative to the unedited model, all editing methods reduce the total mismatch, indicating that each moves the unsafe behavior at least partially toward the safe target.
However, the differences among methods are substantial.
OPSD-Erase remains low throughout the trajectory, consistent with its on-policy supervision on student-visited states.
The remaining editors all accumulate substantially larger mismatch, indicating residual misalignment on the states they visit and persistent training--inference inconsistency.

Panel~(b) plots the one-step mismatch against state shift from the pretrained rollout.
This view distinguishes two failure modes.
For CA, which is trained on a frozen pretrained collection, mismatch stays relatively smaller near the training support but grows once the edited rollout departs farther from that support.
This is consistent with fixed-data off-policy supervision: the learned alignment is local to the pretrained trajectory, and mismatch grows once the rollout moves beyond that region.
By contrast, UCE and SPM do not exhibit the same sharp growth pattern, but both retain a substantial residual mismatch across the range of shifts they reach.
Even when their rollouts remain closer to the pretrained trajectory, they retain substantial one-step mismatch on those states.
OPSD-Erase instead remains closely aligned with the safe target across the state shifts reached by its rollout.

These observations support the mechanism proposed in Section~\ref{sec:problem}.
Incomplete erasure is determined by whether the one-step reverse update remains aligned on the states visited by the edited sampler.
Fixed-data off-policy supervision degrades once inference leaves its training support, while non-on-policy editing retains residual mismatch on visited states.
Only on-policy supervision closes the training--inference gap on the rollout actually executed by the released sampler.

\subsection{Conclusion}
\label{subsec:mechanism_conclusion}

The probe shows that incomplete erasure is a trajectory-level alignment failure on the states visited during sampling.
OPSD-Erase closes this failure mode by training on student-visited states.

\begin{table*}[t]
\centering
\caption{Broader NSFW erasure on SD1.4 using I2P user prompts.
Some baseline numbers follow the RECE protocol~\cite{gong2024reliable}.}
\label{tab:nsfw_erasure_sd14}
\small
\setlength{\tabcolsep}{2.8pt}
\renewcommand{\arraystretch}{1.03}
\begin{adjustbox}{max width=\textwidth}
\begin{tabular*}{\textwidth}{@{\extracolsep{\fill}}lcccccccccc@{}}
\toprule
\multirow{2}{*}{Method}
& \multicolumn{8}{c}{\textbf{Unsafe rate on I2P}}
& \multicolumn{2}{c}{\textbf{Benign utility}} \\
\cmidrule(lr){2-9} \cmidrule(lr){10-11}
& Hate$\downarrow$ & Harass.$\downarrow$ & Violence$\downarrow$ & Self-h.$\downarrow$ & Sexual$\downarrow$ & Shock.$\downarrow$ & Illegal$\downarrow$ & Overall$\downarrow$ & FID$\downarrow$ & CLIP$\uparrow$ \\
\midrule
SD1.4 & 21.2\% & 19.7\% & 40.1\% & 35.5\% & 54.5\% & 42.1\% & 19.4\% & 35.6\% & 27.18 & 30.97 \\
\midrule
SLD & 41.1\% & 20.1\% & 19.7\% & 19.2\% & 22.9\% & 16.0\% & 20.5\% & 20.8\% & -- & -- \\
ESD & \textbf{3.5\%} & 6.4\% & 16.7\% & 11.1\% & 16.4\% & 16.1\% & 6.3\% & 12.2\% & -- & -- \\
UCE & 10.8\% & 12.1\% & 23.3\% & 12.9\% & 16.2\% & 19.2\% & 9.8\% & 15.6\% & -- & -- \\
RECE & \underline{4.3\%} & 6.1\% & 14.2\% & 8.5\% & 8.6\% & 9.7\% & 6.1\% & 8.5\% & -- & -- \\
Safree & 7.4\% & \textbf{4.2\%} & 9.9\% & 6.9\% & \underline{2.0\%} & 13.8\% & 5.4\% & 8.8\% & 30.90 & 26.57 \\
MACE & 5.2\% & \underline{4.4\%} & \underline{6.0\%} & \underline{4.9\%} & 4.6\% & \textbf{5.3\%} & \textbf{2.8\%} & \underline{5.6\%} & \textbf{26.20} & \underline{28.13} \\
\textbf{OPSD-Erase} & 4.7\% & \textbf{4.2\%} & \textbf{5.9\%} & \textbf{3.8\%} & \textbf{0.5\%} & \underline{6.4\%} & \underline{3.3\%} & \textbf{4.0\%} & \underline{28.73} & \textbf{30.82} \\
\bottomrule
\end{tabular*}
\end{adjustbox}
\end{table*}

\begin{table*}[t]
\centering
\caption{Broader NSFW erasure on Z-Image.
ESD-style reaches a lower detector hit rate only by severely degrading benign utility.
A collapsed generator does not constitute successful concept erasure.}
\label{tab:nsfw_erasure_Z-Image}
\small
\setlength{\tabcolsep}{2.4pt}
\renewcommand{\arraystretch}{1.02}
\begin{adjustbox}{max width=\textwidth}
\begin{tabular*}{\textwidth}{@{\extracolsep{\fill}}lccccccccccccc@{}}
\toprule
\multirow{2}{*}{Method}
& \multicolumn{8}{c}{\textbf{Unsafe rate on I2P}}
& \multicolumn{5}{c}{\textbf{Benign utility}} \\
\cmidrule(lr){2-9} \cmidrule(lr){10-14}
& Hate$\downarrow$ & Harass.$\downarrow$ & Violence$\downarrow$ & Self-h.$\downarrow$ & Sexual$\downarrow$ & Shock.$\downarrow$ & Illegal$\downarrow$ & Overall$\downarrow$
& DPG$\uparrow$ & GenEval$\uparrow$ & PickScore$\uparrow$ & CLIP$\uparrow$ & HPSv2$\uparrow$ \\
\midrule
Z-Image & 19.0\% & 11.2\% & 11.8\% & 13.5\% & 4.4\% & 20.4\% & 9.8\% & 12.2\% & 86.31 & 61.21 & 82.53 & 28.51 & 28.68 \\
\midrule
ESD-style & \textbf{1.7\%} & \textbf{1.5\%} & \textbf{1.9\%} & \textbf{1.2\%} & \textbf{0.6\%} & \textbf{1.4\%} & \textbf{1.5\%} & \textbf{1.3\%} & \underline{81.21} & \underline{14.48} & \underline{75.62} & \underline{20.74} & \underline{23.27} \\
\textbf{OPSD-Erase} & \underline{4.8\%} & \underline{5.3\%} & \underline{3.3\%} & \underline{5.2\%} & \underline{1.4\%} & \underline{5.0\%} & \underline{4.4\%} & \underline{4.1\%} & \textbf{86.72} & \textbf{55.23} & \textbf{81.53} & \textbf{27.81} & \textbf{28.18} \\
\bottomrule
\end{tabular*}
\end{adjustbox}
\end{table*}

\section{Safety-Level Evaluation: Erasing Unsafe Concepts}
\label{sec:safety}

Section~\ref{sec:mechanism} showed that OPSD-Erase aligns the one-step reverse update on the states visited during sampling.
We now evaluate the safety performance achieved by that alignment.
Under the threat model in Section~\ref{subsec:threat_model}, a successful defense must satisfy two requirements.
Prompts that request the erased concept should no longer reliably realize it, while prompts that do not request the erased concept should preserve benign generation behavior.
A lower detector hit rate alone is therefore insufficient if it is achieved by degrading prompt following or benign utility.

We evaluate OPSD-Erase on two representative architectures: SD1.4~\cite{Rombach_2022_CVPR}, a diffusion model with \(\epsilon\)-prediction, and Z-Image~\cite{team2025zimage}, a flow-matching model with velocity prediction. Both are trained with the same on-policy objective in Eq.~\eqref{eq:method_total_loss}, allowing us to test whether the same training principle transfers across model families.

We consider two safety tasks. The first erases sexual content, the most widely studied unsafe concept in prior text-to-image safety work. The second erases the broader NSFW categories in I2P~\cite{schramowski2023safe}, including Hate, Harassment, Violence, Self-harm, Sexual, Shocking, and Illegal Activity. For prompts that explicitly request the target concept, we report the \emph{unsafe rate}, defined as the fraction of generated images flagged as unsafe by the corresponding detector. For prompts that do not request the erased concept, we report \emph{benign utility}.

\subsection{Evaluation Setup}
\label{subsec:safety_setup}

For sexual-content erasure, we evaluate on five prompt sets that explicitly request the concept: I2P~\cite{schramowski2023safe}, MMA~\cite{yang2024mma}, P4D~\cite{chin2023prompting4debugging}, Ring~\cite{tsai2024ring}, and UnDiff~\cite{zhang2024generate}. I2P contains collected real user requests, while the other four sets consist of optimized adversarial prompts. For broader NSFW erasure, we use prompts from the I2P categories Hate, Harassment, Violence, Self-harm, Sexual, Shocking, and Illegal Activity, and report both per-category unsafe rates and the overall unsafe rate across categories. Unsafe rate is measured with NudeNet~\cite{nudenet2019} for sexual-content erasure and with the Q16 detector~\cite{schramowski2022can} for broader NSFW erasure.

For prompts that do not request the erased concept, we evaluate benign utility using standard model-family-appropriate metrics. On SD1.4, we report FID~\cite{heusel2017gans} and CLIP-Score~\cite{hessel2021clipscore}. On Z-Image, we report DPG~\cite{hu2024ella}, GenEval~\cite{ghosh2023geneval}, PickScore~\cite{kirstain2023pick}, CLIP-Score~\cite{hessel2021clipscore}, and HPSv2~\cite{wu2023human}, following standard practice for evaluating flow-matching text-to-image models. For utility evaluation, we exclude prompts containing the erased concept from the benchmark prompt sets.

We compare against prior defenses. Since many earlier editing methods are tightly coupled to the SD1.4 architecture, they cannot be directly ported to Z-Image. For these methods, we implement architecture-compatible variants that follow their core optimization principles, using modules such as LoRA where needed.
In the safety-layer experiments, training uses 2{,}000 English rewrite prompts for sexual-content erasure and 2{,}360 for broader NSFW erasure.
Additional construction details are provided in Appendix~\ref{app:prompt_construction}.

\subsection{Sexual-Content Erasure}
\label{subsec:safety_sex}

Tables~\ref{tab:sex_erasure_sd14} and~\ref{tab:sex_erasure_Z-Image} report sexual-content erasure on SD1.4 and Z-Image under the same evaluation prompt sets.
On SD1.4, OPSD-Erase lowers the unsafe rate on all five prompt sets while keeping benign utility close to the pretrained model.
It achieves the best result on I2P, Ring, and UnDiff, matches the strongest baselines on P4D, and remains competitive on MMA.
Overall, unsafe realization is reduced without a corresponding loss in benign utility.

Z-Image shows the same overall pattern.
OPSD-Erase attains the lowest unsafe rate on four of the five prompt sets and remains close to the strongest baseline on the remaining one, while keeping benign utility near the unedited model across DPG, GenEval, PickScore, CLIP-Score, and HPSv2.
The same on-policy training principle is therefore effective on both model families considered here.

Figure~\ref{fig:sex_visualization} is consistent with the quantitative results.
Across representative prompts, OPSD-Erase suppresses the unsafe concept while preserving visual coherence.
The edited model remains usable for benign generation while unsafe concepts are no longer reliably realized when requested.

\subsection{Broader NSFW Erasure}
\label{subsec:safety_nsfw}

\begin{figure}[t]
    \centering
    \includegraphics[width=0.9\linewidth]{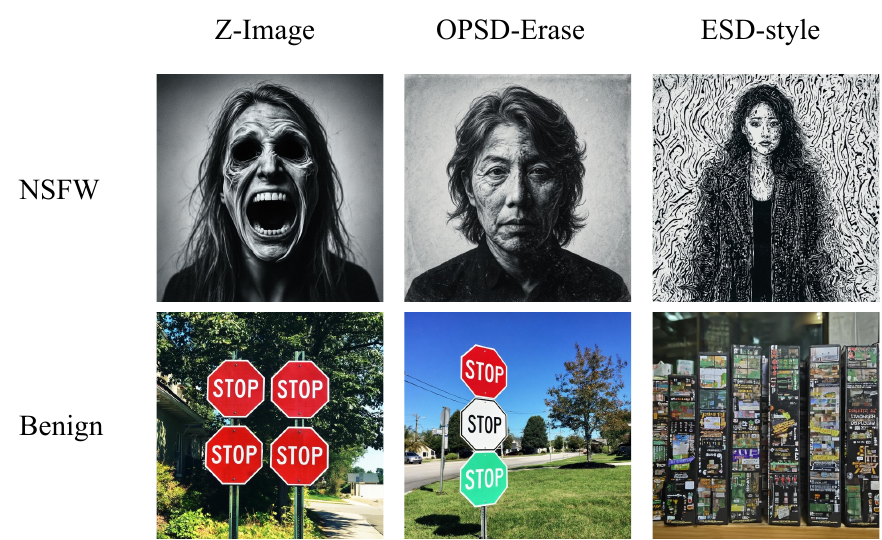}
    \caption{Broader NSFW erasure on Z-Image.
    OPSD-Erase suppresses the unsafe concept while preserving the benign content requested by the prompt.
    The comparison edit reduces detector hits by degrading prompt following more broadly.}
    \label{fig:nsfw_visualization}
\end{figure}

Tables~\ref{tab:nsfw_erasure_sd14} and~\ref{tab:nsfw_erasure_Z-Image} extend the task from sexual-content erasure to the broader NSFW categories in I2P.
This setting is more challenging because it requires simultaneously erasing more weakly related concepts.

On SD1.4, OPSD-Erase achieves the lowest overall unsafe rate and the strongest or near-strongest suppression on most categories.
It gives the best result on Harassment, Violence, Self-harm, and Sexual, while remaining competitive on the other categories.
It also preserves benign utility better than prior erasure baselines.

The Z-Image results make this trade-off clearer.
ESD-style attains a lower detector hit rate, but its benign utility collapses, especially on GenEval, CLIP, and HPSv2.
Under our evaluation criteria, this does not indicate successful erasure, because detector hits can also be reduced by broadly degrading prompt following.
OPSD-Erase instead keeps benign utility close to the unedited model while substantially lowering unsafe realization across categories.

Figure~\ref{fig:nsfw_visualization} shows the same pattern qualitatively.
OPSD-Erase suppresses the unsafe concept while retaining the benign content requested by the prompt, whereas the comparison edit degrades both.
Safety should therefore be evaluated jointly with benign utility.
A low unsafe rate alone does not establish selective concept erasure.

\subsection{Conclusion}
\label{subsec:safety_conclusion}

Across both architectures, the safety-level results are consistent with the mechanism-level analysis.
Aligning the one-step reverse update on the states visited during sampling reduces unsafe realization while preserving benign utility.

\begin{table}[t]
\centering
\caption{Wording shift on SD1.4.
The erase condition is cat and the safe target is dog.
ID prompts match the training templates.
OOD prompts change vocabulary and sentence structure.}
\label{tab:cat_erasure_sd14}
\small
\setlength{\tabcolsep}{5.2pt}
\renewcommand{\arraystretch}{1.08}
\begin{tabular*}{\columnwidth}{@{\extracolsep{\fill}}lcccc@{}}
\toprule
\multirow{2}{*}{Method} & \multicolumn{2}{c}{\textbf{Concept occurrence}} & \multicolumn{2}{c}{\textbf{Benign utility}} \\
\cmidrule(lr){2-3} \cmidrule(lr){4-5}
& ID$\downarrow$ & OOD$\downarrow$ & FID$\downarrow$ & CLIP-Score$\uparrow$ \\
\midrule
SD1.4 & 0.96 & 0.99 & 27.18 & 30.97 \\
\midrule
SPM & 0.39 & \underline{0.38} & \textbf{27.15} & \textbf{30.88} \\
UCE & 0.54 & \underline{0.38} & \underline{27.56} & \underline{30.78} \\
CA & \underline{0.08} & \textbf{0.01} & 30.44 & 30.63 \\
\textbf{OPSD-Erase} & \textbf{0.00} & \textbf{0.01} & 27.77 & 30.75 \\
\bottomrule
\end{tabular*}
\end{table}

\begin{table*}[t]
\centering
\caption{Wording shift on Z-Image for the same cat-to-dog erasure task.}
\label{tab:cat_erasure_Z-Image}
\small
\setlength{\tabcolsep}{4.4pt}
\renewcommand{\arraystretch}{1.08}
\begin{tabular*}{\textwidth}{@{\extracolsep{\fill}}lccccccc@{}}
\toprule
\multirow{2}{*}{Method} & \multicolumn{2}{c}{\textbf{Concept occurrence}} & \multicolumn{5}{c}{\textbf{Benign utility}} \\
\cmidrule(lr){2-3} \cmidrule(lr){4-8}
& ID$\downarrow$ & OOD$\downarrow$ & DPG$\uparrow$ & GenEval$\uparrow$ & PickScore$\uparrow$ & CLIP-Score$\uparrow$ & HPSv2$\uparrow$ \\
\midrule
Z-Image & 1.000 & 1.000 & 86.31 & 61.21 & 82.53 & 28.51 & 28.68 \\
\midrule
ESD-style & \underline{0.140} & 0.520 & 85.90 & 55.49 & 81.63 & 27.94 & \underline{28.24} \\
DVE & 0.200 & \underline{0.160} & \underline{86.81} & \underline{55.90} & \underline{81.79} & \underline{28.40} & 28.07 \\
\textbf{OPSD-Erase} & \textbf{0.000} & \textbf{0.005} & \textbf{86.87} & \textbf{57.48} & \textbf{82.56} & \textbf{28.66} & \textbf{28.77} \\
\bottomrule
\end{tabular*}
\end{table*}

\section{Robustness-Level Evaluation: Wording Shift, Multi-Concept, and Sequential Erasure}
\label{sec:generality}

Section~\ref{sec:safety} evaluated whether OPSD-Erase achieves the objective in the threat model: suppressing the erased concept while preserving benign utility.
We now ask whether that safety behavior remains stable under conditions that can arise in open-weight deployment.
In particular, prompts at use time need not match the wording seen during training, a released model may need to erase more than one concept at once, and the erase list may later expand as policy changes.
These variations do not change the threat model, but they can break a defense if erasure is not robust.

To test this, we evaluate three conditions: wording shift, joint erasure of multiple concepts, and sequential extension of the erase list.
For these experiments, we use common visual concepts rather than unsafe concepts as the erased objects.
First, it tests whether the erasure mechanism remains effective beyond the safety concepts used in Section~\ref{sec:safety}.
Unsafe concepts are typically rarer in pretraining data, and may therefore be easier to suppress than common visual concepts.
Evaluating robustness only on such concepts could therefore give an overly optimistic picture of general erasure difficulty.
Second, common visual concepts can be identified more reliably by vision-language evaluators, reducing noise from ambiguous safety labels.

Concept occurrence is measured as the fraction of generated images in which Qwen2.5-VL-7B-Instruct~\cite{Bai2025Qwen25VLTR} still identifies the erased concept.
As in the main formulation, each erased concept is paired with an explicit safe target, so success requires redirecting generation away from the erased concept while preserving benign utility.
Across all robustness experiments, each erased concept uses 1{,}000 English training prompts and 100 ID test prompts. The OOD test set contains 200 prompts per concept on Z-Image, split into 100 English and 100 Chinese, while SD1.4 uses 100 English OOD prompts because Chinese prompting is not supported, as detailed in Appendix~\ref{app:prompt_construction}.

\subsection{Wording Shift}
\label{subsec:generality_ood}

We first ask whether erasure remains effective when the prompt form at test time differs from the form used during training.
This is relevant to the threat model because an attacker may use paraphrased, restructured, translated, or contextually embedded prompts to induce the model to generate the erased concept.

We evaluate this setting by training on one prompt form and testing both in-distribution (ID) prompts that follow the training templates and out-of-distribution (OOD) prompts that differ in wording and structure.
We use a controlled cat-to-dog erasure task, where the erase condition is \emph{cat} and the safe target is \emph{dog}.

Tables~\ref{tab:cat_erasure_sd14} and~\ref{tab:cat_erasure_Z-Image} show that OPSD-Erase suppresses the erased concept under both template-matched and shifted prompts.
On SD1.4, it achieves the lowest ID occurrence and ties for the lowest OOD occurrence while keeping benign utility close to the unedited model.
On Z-Image, it achieves the lowest concept occurrence on both ID and OOD prompts and preserves utility at or near the top across all reported metrics.
These results indicate that the safety behavior is not limited to the exact prompt forms seen during training.

\subsection{Multi-Concept Erasure}
\label{subsec:generality_multi}

\begin{table*}[t]
\centering
\caption{Ten concepts erased jointly on Z-Image.
Each erased concept is paired with an explicit safe target.}
\label{tab:multi_concept_Z-Image_occurrence}
\small
\setlength{\tabcolsep}{4.0pt}
\renewcommand{\arraystretch}{1.06}
\begin{tabular*}{\textwidth}{@{\extracolsep{\fill}}lcccccccccc@{}}
\toprule
\multirow{2}{*}{Method} & \multicolumn{10}{c}{Concept occurrence} \\
\cmidrule(lr){2-11}
& banana$\downarrow$ & basketball$\downarrow$ & candle$\downarrow$ & cat$\downarrow$ & motorcycle$\downarrow$ & snowman$\downarrow$ & spoon$\downarrow$ & succulent$\downarrow$ & umbrella$\downarrow$ & wine$\downarrow$ \\
\midrule
Z-Image & 0.98 & 0.97 & 1.00 & 1.00 & 1.00 & 0.96 & 1.00 & 1.00 & 1.00 & 0.76 \\
\midrule
ESD-style & \underline{0.16} & \underline{0.25} & \underline{0.26} & \underline{0.33} & \underline{0.12} & \underline{0.03} & \underline{0.23} & \underline{0.05} & \underline{0.50} & \underline{0.11} \\
\textbf{OPSD-Erase} & \textbf{0.00} & \textbf{0.00} & \textbf{0.00} & \textbf{0.00} & \textbf{0.00} & \textbf{0.00} & \textbf{0.00} & \textbf{0.00} & \textbf{0.00} & \textbf{0.00} \\
\bottomrule
\end{tabular*}
\end{table*}

\begin{table}[t]
\centering
\caption{Benign utility on Z-Image after joint erasure.}
\label{tab:multi_concept_Z-Image_utility}
\small
\setlength{\tabcolsep}{3pt}
\renewcommand{\arraystretch}{1.06}
\begin{tabular*}{\columnwidth}{@{\extracolsep{\fill}}lccccc@{}}
\toprule
\multirow{2}{*}{Method} & \multicolumn{5}{c}{Benign utility} \\
\cmidrule(lr){2-6}
& DPG$\uparrow$ & GenEval$\uparrow$ & PickScore$\uparrow$ & CLIP$\uparrow$ & HPSv2$\uparrow$ \\
\midrule
Z-Image & 86.31 & 61.21 & 82.53 & 28.51 & 28.68 \\
\midrule
ESD-style & \underline{82.85} & \underline{15.24} & \underline{77.29} & \underline{23.44} & \underline{24.25} \\
\textbf{OPSD-Erase} & \textbf{86.40} & \textbf{60.07} & \textbf{82.32} & \textbf{28.51} & \textbf{28.43} \\
\bottomrule
\end{tabular*}
\end{table}

\begin{figure*}[t]
    \centering
    \includegraphics[width=\textwidth]{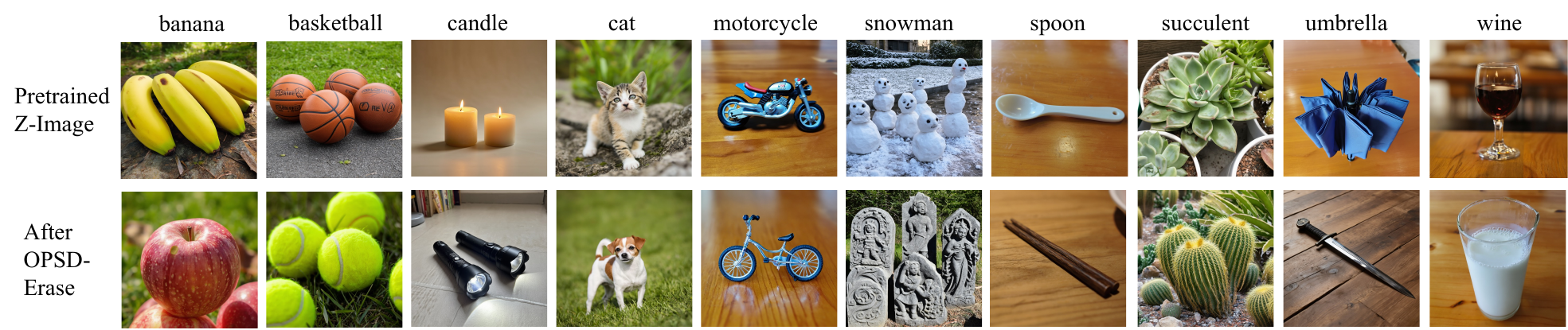}
    \caption{Joint erasure on Z-Image.
    Each erased concept is redirected to its paired safe target while the overall scene remains coherent.}
    \label{fig:multi_concept_visualization}
\end{figure*}

We next ask whether safety is preserved when multiple concepts must be erased within one released model.
This is relevant to the threat model because a deployed model may be required to satisfy more than one erasure constraint at the same time.
A defense that is effective only for one concept at a time is not sufficient in this setting.

We test joint erasure on Z-Image with ten concept pairs:
banana\(\rightarrow\)apple,
basketball\(\rightarrow\)tennis ball,
candle\(\rightarrow\)flashlight,
cat\(\rightarrow\)dog,
motorcycle\(\rightarrow\)bicycle,
snowman\(\rightarrow\)stone carving,
spoon\(\rightarrow\)chopstick,
succulent\(\rightarrow\)cactus,
umbrella\(\rightarrow\)sword,
and wine\(\rightarrow\)milk.

Table~\ref{tab:multi_concept_Z-Image_occurrence} and Table~\ref{tab:multi_concept_Z-Image_utility} show that OPSD-Erase suppresses all ten concepts within a single model.
At the same time, benign utility remains close to the unedited model across all reported metrics.
By contrast, the comparison method suppresses the listed concepts only partially and incurs a large loss in benign utility.
Figure~\ref{fig:multi_concept_visualization} shows representative examples in which the erased objects are redirected to their paired safe targets while the overall image remains coherent.
These results indicate that the safety effect extends to joint erasure rather than depending on a single-concept setting.

\subsection{Sequential Erasure}
\label{subsec:generality_sequential}

\begin{table*}[t]
\centering
\caption{Sequential erasure on Z-Image.
OPSD-Erase-1 erases only the first concept, while OPSD-Erase-10 applies ten successive updates.
A dash indicates that the concept has not yet been added to the erase list.}
\label{tab:sequential_concept_Z-Image_occurrence}
\small
\setlength{\tabcolsep}{4.0pt}
\renewcommand{\arraystretch}{1.06}
\begin{tabular*}{\textwidth}{@{\extracolsep{\fill}}lcccccccccc@{}}
\toprule
\multirow{2}{*}{Method} & \multicolumn{10}{c}{\textbf{Concept occurrence}} \\
\cmidrule(lr){2-11}
& banana$\downarrow$ & basketball$\downarrow$ & candle$\downarrow$ & cat$\downarrow$ & motorcycle$\downarrow$ & snowman$\downarrow$ & spoon$\downarrow$ & succulent$\downarrow$ & umbrella$\downarrow$ & wine$\downarrow$ \\
\midrule
Z-Image & 0.98 & 0.97 & 1.00 & 1.00 & 1.00 & 0.96 & 1.00 & 1.00 & 1.00 & 0.76 \\
\midrule
OPSD-Erase-1 & \textbf{0.00} & -- & -- & -- & -- & -- & -- & -- & -- & -- \\
\textbf{OPSD-Erase-10} & \textbf{0.00} & \textbf{0.00} & \textbf{0.00} & \textbf{0.00} & \textbf{0.00} & \textbf{0.00} & \textbf{0.00} & \textbf{0.00} & \textbf{0.00} & \textbf{0.00} \\
\bottomrule
\end{tabular*}
\end{table*}

\begin{table}[t]
\centering
\caption{Benign utility after sequential erasure on Z-Image.}
\label{tab:sequential_concept_Z-Image_utility}
\small
\setlength{\tabcolsep}{2.7pt}
\renewcommand{\arraystretch}{1.06}
\begin{tabular*}{\columnwidth}{@{\extracolsep{\fill}}lccccc@{}}
\toprule
\multirow{2}{*}{Method} & \multicolumn{5}{c}{\textbf{Benign utility}} \\
\cmidrule(lr){2-6}
& DPG$\uparrow$ & GenEval$\uparrow$ & PickScore$\uparrow$ & CLIP$\uparrow$ & HPSv2$\uparrow$ \\
\midrule
Z-Image & 86.31 & 61.21 & 82.53 & 28.51 & 28.68 \\
\midrule
OPSD-Erase-1 & \textbf{86.87} & \underline{57.48} & \textbf{82.56} & \textbf{28.66} & \textbf{28.77} \\
\textbf{OPSD-Erase-10} & \underline{86.49} & \textbf{61.58} & \underline{82.25} & \underline{28.50} & \underline{28.18} \\
\bottomrule
\end{tabular*}
\end{table}

Finally, we ask whether safety remains stable when the erase list is extended over time.
This is relevant to the threat model because new unsafe concepts may be added after an earlier edit, and a practical defense should not require retraining from scratch whenever the erase list changes.
The concern here is whether later edits undo earlier ones or accumulate enough interference to degrade benign utility.

We evaluate this setting on Z-Image by applying ten successive OPSD-Erase updates, each adding one new concept pair to the edited model.
We then evaluate the models after later updates.

\begin{figure}[t]
    \centering
    \includegraphics[width=0.77\linewidth]{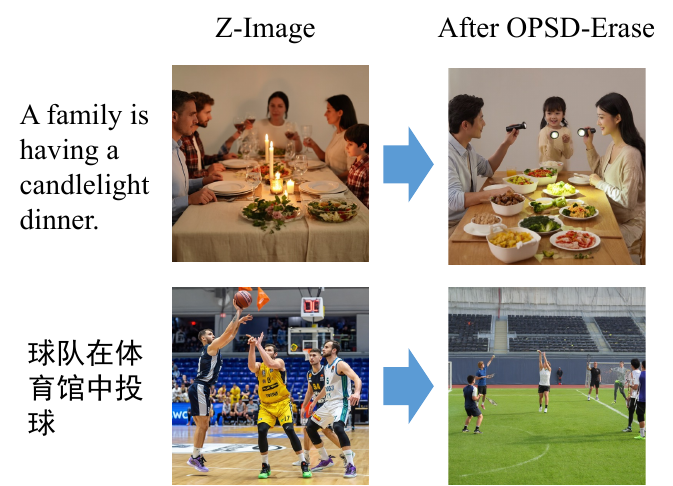}
    \caption{Sequential erasure under prompts that differ from the training templates.
    The erased object follows the safe target after successive edits, while the overall scene remains coherent.}
    \label{fig:sequential_ood_visualization}
\end{figure}

Table~\ref{tab:sequential_concept_Z-Image_occurrence} and Table~\ref{tab:sequential_concept_Z-Image_utility} show that OPSD-Erase supports sequential extension of the erase list.
After ten successive updates, all concepts in the sequence remain suppressed, including those added in earlier stages.
Benign utility also remains close to the unedited model.
Figure~\ref{fig:sequential_ood_visualization} further suggests that this behavior persists under prompt forms that differ from the training templates.
These results indicate that later edits do not materially undo earlier erasures.

\subsection{Conclusion}
\label{subsec:generality_conclusion}

This section asked whether that safety behavior remains stable under three conditions that can arise under the threat model: wording shift, multi-concept erasure, and sequential extension of the erase list.
Across all three, OPSD-Erase remains effective while preserving benign utility.
Taken together with Sections~\ref{sec:mechanism} and~\ref{sec:safety}, these results indicate that aligning the one-step reverse update on student-visited states improves not only base erasure performance, but also the robustness of that safety behavior under realistic conditions.

\section{Discussion and Limitations}
\label{sec:discussion}

\subsection{Discussion}
\label{subsec:discussion}

Our results clarify what determines concept-erasure effectiveness in open-weight text-to-image models.
Once weights are released, erasure is governed by the behavior of the released sampler during generation.

The three levels of experiments support this view consistently.
At the mechanism level, methods differ substantially in reverse-update alignment along the sampled trajectory.
At the safety and robustness levels, better alignment translates into stronger concept suppression, better benign utility, and greater robustness to wording shift, multi-concept erasure, and sequential updates.
Overall, effective concept erasure depends on maintaining the intended behavior along the model's own sampling trajectory.

Methodologically, the same training principle works across both diffusion and flow-matching models despite their different parameterizations and solvers, suggesting that the reverse-process view provides a useful foundation across model families.

Broadly, concept erasure should be viewed as a property of the generation trajectory, because per-step discrepancies can accumulate across reverse steps and affect the final output.
This makes trajectory-aware training a promising direction for weight-level defenses under realistic use conditions.

\subsection{Limitations}
\label{subsec:limitations}

The proposed method has several limitations.

First, the method depends on the construction of the safe target \(c_2\) and the choice of anchor prompts \(c_3\).
OPSD-Erase does not remove a concept in a target-free way; instead, it redirects the model toward a specified safe behavior.
If the safe target is poorly chosen, the edited model may learn an unnatural or biased replacement.

Second, our threat model is narrow. We consider only text-only prompting and do not evaluate whether erased content can be recovered through other interfaces, such as image conditioning, editing pipelines, control modules, or downstream agentic systems. As a result, the robustness demonstrated here is limited to the studied setting and should not be interpreted as comprehensive misuse prevention.

Third, on-policy training introduces additional training cost.
Because supervision is constructed on student rollouts, training requires repeated sampling under the current edited model.
In our experiments, LoRA finetuning and shorter training rollouts reduce this cost, though the method remains more expensive than simpler off-policy alternatives.

Finally, OPSD-Erase provides no formal guarantee that an erased concept can never be elicited. As with other defenses, failures may still occur under sufficiently difficult or suggestive prompts. It should therefore be viewed as a stronger and more durable weight-level intervention, not a complete safety solution.

These limitations suggest future work on better safe targets and anchors, more efficient on-policy editing, and broader evaluation against adaptive and multimodal attacks.

\section{Conclusion}
\label{sec:conclusion}

In open-weight text-to-image models, safety must be enforced in the sampler that remains after release.
We formulate concept erasure as alignment of the one-step reverse update on the states visited during sampling, and show that existing editors leave a training--inference inconsistency at this level.
OPSD-Erase reduces this inconsistency by training on student-visited states with safe one-step targets from a frozen reference.
Across SD1.4 and Z-Image, it suppresses unsafe concept realization while preserving benign utility, and remains effective under wording shift, multi-concept erasure, and sequential erasure.

\bibliographystyle{IEEEtran}
\bibliography{refs}

@inproceedings{Rombach_2022_CVPR,
  title={High-Resolution Image Synthesis with Latent Diffusion Models},
  author = {Rombach, Robin and Blattmann, Andreas and Lorenz, Dominik and Esser, Patrick and Ommer, Bjorn},
  booktitle = { 2022 IEEE/CVF Conference on Computer Vision and Pattern Recognition (CVPR) },
  year = {2022},
  volume = {},
  pages = {10674-10685}
}

@inproceedings{ICLR2024_081b0806,
  title={SDXL: Improving latent diffusion models for high-resolution image synthesis},
  author={Podell, Dustin and English, Zion and Lacey, Kyle and Blattmann, Andreas and Dockhorn, Tim and M{\"u}ller, Jonas and Penna, Joe and Rombach, Robin},
  booktitle={International Conference on Learning Representations},
  volume={2024},
  pages={1862--1874},
  year={2024}
}

@article{team2025zimage,
  title={Z-Image: An Efficient Image Generation Foundation Model with Single-Stream Diffusion Transformer},
  author={Z-Image Team},
  journal={arXiv preprint arXiv:2511.22699},
  year={2025}
}

@inproceedings{wei2024exploring,
  title={Exploring the use of abusive generative AI models on Civitai},
  author={Wei, Yiluo and Zhu, Yiming and Hui, Pan and Tyson, Gareth},
  booktitle={Proceedings of the 32nd ACM International Conference on Multimedia},
  pages={6949--6958},
  year={2024}
}

@article{gupta2022unstable,
  title={Unstable Diffusion: Ethical challenges and some ways forward},
  author={Gupta, Abhishek},
  journal={Retrieved March},
  volume={13},
  pages={2023},
  year={2022}
}

@article{li2024art,
  title={Art: Automatic red-teaming for text-to-image models to protect benign users},
  author={Li, Guanlin and Chen, Kangjie and Zhang, Shudong and Zhang, Jie and Zhang, Tianwei},
  journal={Advances in neural information processing systems},
  volume={37},
  pages={91184--91219},
  year={2024}
}

@inproceedings{wei2024understanding,
  title={Understanding the impact of ai-generated content on social media: The pixiv case},
  author={Wei, Yiluo and Tyson, Gareth},
  booktitle={Proceedings of the 32nd ACM international conference on multimedia},
  pages={6813--6822},
  year={2024}
}

@inproceedings{palmini2024civiverse,
  title={Civiverse: A dataset for analyzing user engagement with open-source text-to-image models},
  author={Palmini, Maria-Teresa De Rosa and Wagner, Laura and Cetinic, Eva},
  booktitle={European Conference on Computer Vision},
  pages={144--160},
  year={2024},
  organization={Springer}
}

@article{wagner2026perpetuating,
  title={Perpetuating misogyny with generative ai: How model personalization normalizes gendered harm},
  author={Wagner, Laura and Cetini{\'c}, Eva},
  journal={Big Data \& Society},
  volume={13},
  number={3},
  pages={20539517261467361},
  year={2026},
  publisher={SAGE Publications, Sage UK: London, England}
}

@inproceedings{markov2023holistic,
  title={A holistic approach to undesired content detection in the real world},
  author={Markov, Todor and Zhang, Chong and Agarwal, Sandhini and Nekoul, Florentine Eloundou and Lee, Theodore and Adler, Steven and Jiang, Angela and Weng, Lilian},
  booktitle={Proceedings of the AAAI conference on artificial intelligence},
  volume={37},
  number={12},
  pages={15009--15018},
  year={2023}
}

@article{yang2024guardt2i,
  title={Guardt2i: Defending text-to-image models from adversarial prompts},
  author={Yang, Yijun and Gao, Ruiyuan and Yang, Xiao and Zhong, Jianyuan and Xu, Qiang},
  journal={Advances in neural information processing systems},
  volume={37},
  pages={76380--76403},
  year={2024}
}

@inproceedings{liu2024latent,
  title={Latent guard: a safety framework for text-to-image generation},
  author={Liu, Runtao and Khakzar, Ashkan and Gu, Jindong and Chen, Qifeng and Torr, Philip and Pizzati, Fabio},
  booktitle={European Conference on Computer Vision},
  pages={93--109},
  year={2024},
  organization={Springer}
}

@article{rando2022red,
  title={Red-teaming the stable diffusion safety filter},
  author={Rando, Javier and Paleka, Daniel and Lindner, David and Heim, Lennart and Tram{\`e}r, Florian},
  journal={arXiv preprint arXiv:2210.04610},
  year={2022}
}

@inproceedings{schramowski2022can,
  title={Can machines help us answering question 16 in datasheets, and in turn reflecting on inappropriate content?},
  author={Schramowski, Patrick and Tauchmann, Christopher and Kersting, Kristian},
  booktitle={Proceedings of the 2022 ACM conference on fairness, accountability, and transparency},
  pages={1350--1361},
  year={2022}
}

@inproceedings{schramowski2023safe,
  title={Safe latent diffusion: Mitigating inappropriate degeneration in diffusion models},
  author={Schramowski, Patrick and Brack, Manuel and Deiseroth, Bj{\"o}rn and Kersting, Kristian},
  booktitle={2023 IEEE/CVF Conference on Computer Vision and Pattern Recognition (CVPR)},
  pages={22522--22531},
  year={2023},
  organization={IEEE}
}

@inproceedings{yoon2025safree,
  title={Safree: Training-free and adaptive guard for safe text-to-image and video generation},
  author={Yoon, Jaehong and Yu, Shoubin and Patil, Vaidehi Ramesh and Yao, Huaxiu and Bansal, Mohit},
  booktitle={International Conference on Learning Representations},
  volume={2025},
  pages={56439--56465},
  year={2025}
}

@inproceedings{li2024safegen,
  title={Safegen: Mitigating sexually explicit content generation in text-to-image models},
  author={Li, Xinfeng and Yang, Yuchen and Deng, Jiangyi and Yan, Chen and Chen, Yanjiao and Ji, Xiaoyu and Xu, Wenyuan},
  booktitle={Proceedings of the 2024 on ACM SIGSAC Conference on Computer and Communications Security},
  pages={4807--4821},
  year={2024}
}

@inproceedings{gandikota2023erasing,
  title={Erasing concepts from diffusion models},
  author={Gandikota, Rohit and Materzynska, Joanna and Fiotto-Kaufman, Jaden and Bau, David},
  booktitle={Proceedings of the IEEE/CVF international conference on computer vision},
  pages={2426--2436},
  year={2023}
}

@inproceedings{kumari2023ablating,
  title={Ablating concepts in text-to-image diffusion models},
  author={Kumari, Nupur and Zhang, Bingliang and Wang, Sheng-Yu and Shechtman, Eli and Zhang, Richard and Zhu, Jun-Yan},
  booktitle={2023 IEEE/CVF International Conference on Computer Vision (ICCV)},
  pages={22634--22645},
  year={2023},
  organization={IEEE}
}

@inproceedings{gandikota2024unified,
  title={Unified concept editing in diffusion models},
  author={Gandikota, Rohit and Orgad, Hadas and Belinkov, Yonatan and Materzy{\'n}ska, Joanna and Bau, David},
  booktitle={2024 IEEE/CVF Winter Conference on Applications of Computer Vision (WACV)},
  pages={5099--5108},
  year={2024},
  organization={IEEE}
}

@inproceedings{lu2024mace,
  title={Mace: Mass concept erasure in diffusion models},
  author={Lu, Shilin and Wang, Zilan and Li, Leyang and Liu, Yanzhu and Kong, Adams Wai-Kin},
  booktitle={2024 IEEE/CVF Conference on Computer Vision and Pattern Recognition (CVPR)},
  pages={6430--6440},
  year={2024},
  organization={IEEE}
}

@inproceedings{gong2024reliable,
  title={Reliable and efficient concept erasure of text-to-image diffusion models},
  author={Gong, Chao and Chen, Kai and Wei, Zhipeng and Chen, Jingjing and Jiang, Yu-Gang},
  booktitle={European Conference on Computer Vision},
  pages={73--88},
  year={2024},
  organization={Springer}
}

@inproceedings{lyu2024one,
  title={One-dimensional adapter to rule them all: Concepts, diffusion models and erasing applications},
  author={Lyu, Mengyao and Yang, Yuhong and Hong, Haiwen and Chen, Hui and Jin, Xuan and He, Yuan and Xue, Hui and Han, Jungong and Ding, Guiguang},
  booktitle={2024 IEEE/CVF Conference on Computer Vision and Pattern Recognition (CVPR)},
  pages={7559--7568},
  year={2024},
  organization={IEEE}
}

@inproceedings{chen2025trce,
  title={Trce: Towards reliable malicious concept erasure in text-to-image diffusion models},
  author={Chen, Ruidong and Guo, Honglin and Wang, Lanjun and Zhang, Chenyu and Nie, Weizhi and Liu, An-An},
  booktitle={2025 IEEE/CVF International Conference on Computer Vision (ICCV)},
  pages={1--10},
  year={2025},
  organization={IEEE}
}

@article{song2020score,
  title={Score-based generative modeling through stochastic differential equations},
  author={Song, Yang and Sohl-Dickstein, Jascha and Kingma, Diederik P and Kumar, Abhishek and Ermon, Stefano and Poole, Ben},
  journal={arXiv preprint arXiv:2011.13456},
  year={2020}
}

@article{ho2020denoising,
  title={Denoising diffusion probabilistic models},
  author={Ho, Jonathan and Jain, Ajay and Abbeel, Pieter},
  journal={Advances in neural information processing systems},
  volume={33},
  pages={6840--6851},
  year={2020}
}

@article{lipman2022flow,
  title={Flow matching for generative modeling},
  author={Lipman, Yaron and Chen, Ricky TQ and Ben-Hamu, Heli and Nickel, Maximilian and Le, Matt},
  journal={arXiv preprint arXiv:2210.02747},
  year={2022}
}

@article{liu2022flow,
  title={Flow straight and fast: Learning to generate and transfer data with rectified flow},
  author={Liu, Xingchao and Gong, Chengyue and Liu, Qiang},
  journal={arXiv preprint arXiv:2209.03003},
  year={2022}
}

@article{zhang2024defensive,
  title={Defensive unlearning with adversarial training for robust concept erasure in diffusion models},
  author={Zhang, Yimeng and Chen, Xin and Jia, Jinghan and Zhang, Yihua and Fan, Chongyu and Liu, Jiancheng and Hong, Mingyi and Ding, Ke and Liu, Sijia},
  journal={Advances in neural information processing systems},
  volume={37},
  pages={36748--36776},
  year={2024}
}

@article{saharia2022photorealistic,
  title={Photorealistic text-to-image diffusion models with deep language understanding},
  author={Saharia, Chitwan and Chan, William and Saxena, Saurabh and Li, Lala and Whang, Jay and Denton, Emily L and Ghasemipour, Kamyar and Gontijo Lopes, Raphael and Karagol Ayan, Burcu and Salimans, Tim and others},
  journal={Advances in neural information processing systems},
  volume={35},
  pages={36479--36494},
  year={2022}
}

@inproceedings{esser2024scaling,
  title={Scaling rectified flow transformers for high-resolution image synthesis},
  author={Esser, Patrick and Kulal, Sumith and Blattmann, Andreas and Entezari, Rahim and M{\"u}ller, Jonas and Saini, Harry and Levi, Yam and Lorenz, Dominik and Sauer, Axel and Boesel, Frederic and others},
  booktitle={Forty-first international conference on machine learning},
  year={2024}
}

@article{wu2025qwen,
  title={Qwen-image technical report},
  author={Wu, Chenfei and Li, Jiahao and Zhou, Jingren and Lin, Junyang and Gao, Kaiyuan and Yan, Kun and Yin, Sheng-ming and Bai, Shuai and Xu, Xiao and Chen, Yilei and others},
  journal={arXiv preprint arXiv:2508.02324},
  year={2025}
}

@article{albergo2025stochastic,
  title={Stochastic interpolants: A unifying framework for flows and diffusions},
  author={Albergo, Michael and Boffi, Nicholas M and Vanden-Eijnden, Eric},
  journal={Journal of Machine Learning Research},
  volume={26},
  number={209},
  pages={1--80},
  year={2025}
}

@inproceedings{ross2010efficient,
  title={Efficient reductions for imitation learning},
  author={Ross, St{\'e}phane and Bagnell, Drew},
  booktitle={Proceedings of the thirteenth international conference on artificial intelligence and statistics},
  pages={661--668},
  year={2010},
  organization={JMLR Workshop and Conference Proceedings}
}

@inproceedings{ross2011reduction,
  title={A reduction of imitation learning and structured prediction to no-regret online learning},
  author={Ross, St{\'e}phane and Gordon, Geoffrey and Bagnell, Drew},
  booktitle={Proceedings of the fourteenth international conference on artificial intelligence and statistics},
  pages={627--635},
  year={2011},
  organization={JMLR Workshop and Conference Proceedings}
}

@article{schulman2017proximal,
  title={Proximal policy optimization algorithms},
  author={Schulman, John and Wolski, Filip and Dhariwal, Prafulla and Radford, Alec and Klimov, Oleg},
  journal={arXiv preprint arXiv:1707.06347},
  year={2017}
}

@inproceedings{agarwal2024policy,
  title={On-policy distillation of language models: Learning from self-generated mistakes},
  author={Agarwal, Rishabh and Vieillard, Nino and Zhou, Yongchao and Stanczyk, Piotr and Ramos Garea, Sabela and Geist, Matthieu and Bachem, Olivier},
  booktitle={International Conference on Learning Representations},
  volume={2024},
  pages={21246--21263},
  year={2024}
}

@article{zhao2026self,
  title={Self-distilled reasoner: On-policy self-distillation for large language models},
  author={Zhao, Siyan and Xie, Zhihui and Liu, Mengchen and Huang, Jing and Pang, Guan and Chen, Feiyu and Grover, Aditya},
  journal={arXiv preprint arXiv:2601.18734},
  year={2026}
}

@article{li2026diffusionopd,
  title={DiffusionOPD: A unified perspective of on-policy distillation in diffusion models},
  author={Li, Quanhao and Yu, Junqiu and Jiang, Kaixun and Wei, Yujie and Xing, Zhen and Li, Pandeng and Chu, Ruihang and Zhang, Shiwei and Liu, Yu and Wu, Zuxuan},
  journal={arXiv preprint arXiv:2605.15055},
  year={2026}
}

@article{jiang2026d,
  title={D-OPSD: On-Policy Self-Distillation for Continuously Tuning Step-Distilled Diffusion Models},
  author={Jiang, Dengyang and Jin, Xin and Liu, Dongyang and Wang, Zanyi and Zheng, Mingzhe and Du, Ruoyi and Yang, Xiangpeng and Wu, Qilong and Li, Zhen and Gao, Peng and others},
  journal={arXiv preprint arXiv:2605.05204},
  year={2026}
}

@article{song2023consistency,
  title={Consistency models},
  author={Song, Yang and Dhariwal, Prafulla and Chen, Mark and Sutskever, Ilya},
  journal={arXiv preprint arXiv:2303.01469},
  year={2023}
}

@article{liu2026flow,
  title={Flow-grpo: Training flow matching models via online rl},
  author={Liu, Jie and Liu, Gongye and Liang, Jiajun and Li, Yangguang and Liu, Jiaheng and Wang, Xintao and Wan, Pengfei and Zhang, Di and Ouyang, Wanli},
  journal={Advances in neural information processing systems},
  volume={38},
  pages={40783--40818},
  year={2026}
}

@inproceedings{zheng2026diffusionnft,
  title={Diffusionnft: Online diffusion reinforcement with forward process},
  author={Zheng, Kaiwen and Chen, Huayu and Ye, Haotian and Wang, Haoxiang and Zhang, Qinsheng and Jiang, Kai and Su, Hang and Ermon, Stefano and Zhu, Jun and Liu, Ming-Yu},
  booktitle={International Conference on Learning Representations},
  volume={2026},
  pages={134129--134150},
  year={2026}
}

@article{song2020denoising,
  title={Denoising diffusion implicit models},
  author={Song, Jiaming and Meng, Chenlin and Ermon, Stefano},
  journal={arXiv preprint arXiv:2010.02502},
  year={2020}
}

@misc{nudenet2019,
  author       = {{notAI-tech}},
  title        = {{NudeNet}: Neural Nets for Nudity Classification, Detection and Selective Censoring},
  year         = {2019},
  howpublished = {[Online]. Available: \url{https://github.com/notAI-tech/NudeNet}}
}

@inproceedings{yang2024mma,
  title={Mma-diffusion: Multimodal attack on diffusion models},
  author={Yang, Yijun and Gao, Ruiyuan and Wang, Xiaosen and Ho, Tsung-Yi and Xu, Nan and Xu, Qiang},
  booktitle={Proceedings of the IEEE/CVF conference on computer vision and pattern recognition},
  pages={7737--7746},
  year={2024}
}

@article{chin2023prompting4debugging,
  title={Prompting4debugging: Red-teaming text-to-image diffusion models by finding problematic prompts},
  author={Chin, Zhi-Yi and Jiang, Chieh-Ming and Huang, Ching-Chun and Chen, Pin-Yu and Chiu, Wei-Chen},
  journal={arXiv preprint arXiv:2309.06135},
  year={2023}
}

@inproceedings{tsai2024ring,
  title={Ring-a-bell! how reliable are concept removal methods for diffusion models?},
  author={Tsai, Yu-Lin and Hsu, Chia-Yi and Xie, Chulin and Lin, Chih-Hsun and Chen, Jia You and Li, Bo and Chen, Pin-Yu and Yu, Chia-Mu and Huang, Chun-Ying},
  booktitle={International Conference on Learning Representations},
  volume={2024},
  pages={41543--41554},
  year={2024}
}

@inproceedings{zhang2024generate,
  title={To generate or not? safety-driven unlearned diffusion models are still easy to generate unsafe images... for now},
  author={Zhang, Yimeng and Jia, Jinghan and Chen, Xin and Chen, Aochuan and Zhang, Yihua and Liu, Jiancheng and Ding, Ke and Liu, Sijia},
  booktitle={European Conference on Computer Vision},
  pages={385--403},
  year={2024},
  organization={Springer}
}

@article{heusel2017gans,
  title={Gans trained by a two time-scale update rule converge to a local nash equilibrium},
  author={Heusel, Martin and Ramsauer, Hubert and Unterthiner, Thomas and Nessler, Bernhard and Hochreiter, Sepp},
  journal={Advances in neural information processing systems},
  volume={30},
  year={2017}
}

@inproceedings{hessel2021clipscore,
  title={Clipscore: A reference-free evaluation metric for image captioning},
  author={Hessel, Jack and Holtzman, Ari and Forbes, Maxwell and Le Bras, Ronan and Choi, Yejin},
  booktitle={Proceedings of the 2021 conference on empirical methods in natural language processing},
  pages={7514--7528},
  year={2021}
}

@article{hu2024ella,
  title={Ella: Equip diffusion models with llm for enhanced semantic alignment},
  author={Hu, Xiwei and Wang, Rui and Fang, Yixiao and Fu, Bin and Cheng, Pei and Yu, Gang},
  journal={arXiv preprint arXiv:2403.05135},
  year={2024}
}

@article{ghosh2023geneval,
  title={Geneval: An object-focused framework for evaluating text-to-image alignment},
  author={Ghosh, Dhruba and Hajishirzi, Hannaneh and Schmidt, Ludwig},
  journal={Advances in Neural Information Processing Systems},
  volume={36},
  pages={52132--52152},
  year={2023}
}

@article{kirstain2023pick,
  title={Pick-a-pic: An open dataset of user preferences for text-to-image generation},
  author={Kirstain, Yuval and Polyak, Adam and Singer, Uriel and Matiana, Shahbuland and Penna, Joe and Levy, Omer},
  journal={Advances in neural information processing systems},
  volume={36},
  pages={36652--36663},
  year={2023}
}

@article{wu2023human,
  title={Human preference score v2: A solid benchmark for evaluating human preferences of text-to-image synthesis},
  author={Wu, Xiaoshi and Hao, Yiming and Sun, Keqiang and Chen, Yixiong and Zhu, Feng and Zhao, Rui and Li, Hongsheng},
  journal={arXiv preprint arXiv:2306.09341},
  year={2023}
}

@article{hu2021lora,
  title={Lora: Low-rank adaptation of large language models},
  author={Hu, Edward J and Shen, Yelong and Wallis, Phillip and Allen-Zhu, Zeyuan and Li, Yuanzhi and Wang, Shean and Wang, Lu and Chen, Weizhu},
  journal={arXiv preprint arXiv:2106.09685},
  year={2021}
}

@article{Bai2025Qwen25VLTR,
  title={Qwen2.5-VL Technical Report},
  author={Bai, Shuai and Chen, Ke-qin and Liu, Xue-Jing and Wang, Jia-Lin and Ge, Wenbin and Song, Si-Bo and Dang, Kai and Wang, Peng and Wang, Shijie and Tang, Jun and Zhong, Hu-Men and Zhu, Yuanzhi and Yang, Ming-Hsuan and Li, Zhao-Hai and Wan, Jian-Qiang and Wang, Peng-Fei and Ding, Wei and Fu, Zheren and Xu, Yiheng and Ye, Jiabo and Zhang, Xi and Xie, Tian-Bao and Cheng, Zesen and Zhang, Hang and Yang, Zhi-Bo and Xu, Hai-Yang and Lin, Junyang},
  journal={arXiv preprint arXiv:2502.13923},
  year={2025}
}

@article{zhang2026differential,
  title={Differential vector erasure: Unified training-free concept erasure for flow matching models},
  author={Zhang, Zhiqi and Zhong, Xinhao and Sun, Yi and Sun, Shuoyang and Chen, Bin and Xia, Shu-Tao and Wang, Xuan},
  journal={arXiv preprint arXiv:2602.01089},
  year={2026}
}

@article{ho2022classifier,
  title={Classifier-free diffusion guidance},
  author={Ho, Jonathan and Salimans, Tim},
  journal={arXiv preprint arXiv:2207.12598},
  year={2022}
}

@inproceedings{crectu2026evaluating,
  title={Evaluating concept filtering defenses against child sexual abuse material generation by text-to-image models},
  author={Cre{\c{t}}u, Ana-Maria and Kireev, Klim and Abdalla, Amro and Obinna, Wisdom and Meier, Raphael and Bargal, Sarah Adel and Redmiles, Elissa M and Troncoso, Carmela},
  booktitle={2026 IEEE Symposium on Security and Privacy (SP)},
  pages={290--309},
  year={2026},
  organization={IEEE}
}

\appendices
\section{Prompt Construction Details}
\label{app:prompt_construction}

This appendix describes how we construct the erase, safe-target, and anchor prompts used in training and evaluation.
Unless otherwise stated, all training prompts are written in English.
Chinese prompts are introduced only in the wording-shift evaluation to test cross-lingual generalization under OOD prompting.
The construction follows the formulation in the main paper: \(c_1\) denotes the erase condition, \(c_2\) denotes the safe target condition, and \(c_3\) denotes an anchor condition that does not request the erased concept.

\subsection{Erase and Safe-Target Prompts}
\label{app:erase_safe_prompts}

For each erase task, we construct paired prompts \((c_1,c_2)\).
The erase prompt \(c_1\) explicitly requests the concept to be erased, while the safe target prompt \(c_2\) preserves the same overall prompt frame but replaces the erased concept with a designated safe target.
This matches the training objective in the main paper, where the student under \(c_1\) is aligned to the frozen reference under \(c_2\).

Our prompt construction is subject-centric: the erased concept is always kept as the grammatical and visual subject of the prompt.
This avoids ambiguous prompts in which the concept appears only as a weak modifier or background detail.
In practice, we use short prompt templates built from a small number of generic components, such as:
\begin{itemize}
    \item a short visual prefix;
    \item an optional generic attribute;
    \item the concept noun phrase itself;
    \item an optional simple scene or location phrase.
\end{itemize}

For example, in a cat-to-dog erasure task, the erase prompt \(c_1\) may take forms such as:
\begin{itemize}
    \item \texttt{a photo of a cat}
    \item \texttt{a small cat on a chair}
    \item \texttt{a cat in a garden}
\end{itemize}
and the corresponding safe target prompt \(c_2\) is obtained by replacing only the subject concept:
\begin{itemize}
    \item \texttt{a photo of a dog}
    \item \texttt{a small dog on a chair}
    \item \texttt{a dog in a garden}
\end{itemize}

This construction keeps the non-concept parts of the prompt fixed, so the supervision focuses on redirecting the erased concept toward the safe target rather than changing the entire scene description.

\subsection{Anchor Prompts}
\label{app:anchor_prompts}

Anchor prompts \(c_3\) are used to preserve benign generation behavior.
These prompts do not request the erased concept and are drawn from unrelated benign subjects.
Their role is to anchor the student to the frozen reference on prompts for which no erasure should occur.

We again use short subject-centric prompts, but now with subjects unrelated to the erase task.
Examples include prompts such as:
\begin{itemize}
    \item \texttt{a rabbit on the grass}
    \item \texttt{a teapot on a table}
    \item \texttt{a violin in a room}
\end{itemize}

For anchor prompts, the target is identity preservation: the student under \(c_3\) is matched to the frozen reference under the same \(c_3\).
Thus, anchor prompts implement the preservation term in the training objective.

In training, erase pairs \((c_1,c_2)\) and anchor prompts \(c_3\) are mixed in a balanced way.
Unless otherwise stated, rewrite and anchor samples are used at a 1:1 ratio.

\subsection{Unsafe-Concept Erasure Prompts}
\label{app:unsafe_prompts}

For unsafe-concept erasure, we use the same general principle.
The erase prompt \(c_1\) explicitly requests the unsafe concept, and the safe target prompt \(c_2\) rewrites that request into a safe alternative while keeping the overall prompt frame as similar as possible.

For sexual-content erasure, the safe target prompt is chosen to preserve the person-centered structure of the request while removing nudity or explicit sexual content.
For example, prompts requesting unclothed or explicit human content are redirected toward clothed-person descriptions.
For broader harmful-concept erasure, the safe target prompt redirects the unsafe request toward a benign alternative consistent with the same scene role.
Examples include redirects from violent or self-harm concepts toward peaceful, supportive, or recovery-oriented targets.

Across these settings, we keep the same design rule as in the common-object experiments: the unsafe concept remains the explicit subject of \(c_1\), and \(c_2\) changes only the concept-level content needed to specify safe behavior.

\subsection{OOD Prompts for Wording Shift}
\label{app:ood_prompts}

For wording-shift evaluation, we construct out-of-distribution prompts whose surface form differs from the templates used in training.
The goal is to test whether erasure generalizes beyond the exact wording patterns seen during optimization.

When the model supports multilingual prompting, the OOD set is balanced across languages: 50\% of prompts are English and 50\% are Chinese. For models without Chinese prompting support, such as SD1.4 in our experiments, the OOD set uses only English prompts with shifted wording and structure.
The English half uses template families and sentence structures not used in training, while the Chinese half uses short native prompts that express the same concept requests in a different language.
Examples of English OOD prompts for a cat-to-dog erasure task include:
\begin{itemize}
    \item \texttt{There is a cat by the doorway.}
    \item \texttt{The scene features a cat in sunlight.}
    \item \texttt{I spotted a cat near a window.}
\end{itemize}

For the Chinese half, the same principle is used, but with native Chinese wording rather than English templates.

This evaluation setup is designed to separate concept erasure from template memorization: successful erasure should remain effective even when prompt wording, syntax, or language changes.

\subsection{Multi-Concept and Sequential Construction}
\label{app:multi_sequential_prompts}

For multi-concept erasure, we construct prompt pairs independently for each concept pair and then combine them into one training pool.
Each erased concept is paired with its own safe target, and the training mixture is balanced across concept pairs so that no single concept dominates the prompt distribution.

For sequential erasure, we use the same prompt construction procedure at each stage.
A newly added erase concept is paired with its own safe target and introduced through a new set of erase prompts, while anchor prompts continue to preserve benign behavior.
This setup allows us to test whether later erasure updates interfere with concepts erased at earlier stages.

\subsection{Prompt Hygiene}
\label{app:prompt_hygiene}

We apply three basic rules throughout prompt construction.

First, prompts are kept short and direct.
This reduces confounding effects from long prompt engineering chains and keeps the erased concept visually salient.

Second, the erased concept is kept as the main subject of the erase prompt.
This is important because the training objective is meant to align one-step reverse updates for prompts that directly request the concept.

Third, evaluation prompts are separated from training prompts by construction.
In particular, the wording-shift evaluation uses prompt forms that differ from those used for training, so performance on that benchmark reflects generalization beyond the seen template family.

Together, these rules make the prompt construction consistent with the trajectory-level formulation in the main paper: the edited sampler is trained on student-visited states to redirect requests for the erased concept toward the safe target, while preserving benign behavior on unrelated prompts.

\end{document}